\documentclass{article}

\usepackage[preprint]{neurips_2026}

\usepackage[utf8]{inputenc} % allow utf-8 input
\usepackage[T1]{fontenc}    % use 8-bit T1 fonts
\usepackage{hyperref}       % hyperlinks
\usepackage{url}            % simple URL typesetting
\usepackage{booktabs}       % professional-quality tables
\usepackage{amsfonts}       % blackboard math symbols
\usepackage{nicefrac}       % compact symbols for 1/2, etc.
\usepackage{microtype}      % microtypography
\usepackage{xcolor}         % colors
\usepackage{algorithm}
\usepackage{algpseudocode}
\usepackage{graphicx}
\usepackage{amsmath}
\usepackage{etoolbox}
\AtBeginEnvironment{algorithmic}{\setlength{\itemsep}{0pt}}
\usepackage{wrapfig}      % for wraptable
\usepackage{adjustbox}    % for adjustbox
\usepackage{booktabs}     % for \toprule etc.
\usepackage{nicematrix}   % for NiceTabular
\usepackage{wrapfig}
\usepackage{tikz}

\usepackage{caption}
\usetikzlibrary{arrows.meta, positioning, fit, backgrounds, calc}
\usepackage{tcolorbox}
\usepackage{listings}
\tcbuselibrary{breakable}

\definecolor{pblue}{HTML}{1565C0}
\definecolor{pbluelt}{HTML}{E3F2FD}
\definecolor{pbluemd}{HTML}{BBDEFB}
\definecolor{pred}{HTML}{C62828}
\definecolor{predlt}{HTML}{FFEBEE}
\definecolor{porange}{HTML}{E65100}
\definecolor{porangelt}{HTML}{FFF3E0}
\definecolor{ppurple}{HTML}{6A1B9A}
\definecolor{ppurplelt}{HTML}{F3E5F5}
\definecolor{pgreen}{HTML}{2E7D32}
\definecolor{pgreenlt}{HTML}{E8F5E9}
\definecolor{pgray}{HTML}{546E7A}
\definecolor{pgraylt}{HTML}{ECEFF1}
\definecolor{pagentbg}{HTML}{FFF8E1}
\definecolor{pagentbdr}{HTML}{F9A825}
\definecolor{penvbg}{HTML}{E8F5E9}
\definecolor{penvbdr}{HTML}{66BB6A}

\usepackage{hyperref}
\usepackage[colorinlistoftodos]{todonotes}
\usepackage{colortbl}
\usepackage{nicematrix}
\usepackage{amssymb}
\usepackage{amsmath}
\usepackage{adjustbox}
\usepackage{soul}
\usepackage[inline]{enumitem}
\usepackage{booktabs}
\usepackage{color}
\usepackage{xcolor}
\usepackage{bbding}
\usepackage{listings}
\usepackage{multicol}
\usepackage{xspace}
\usepackage{lmodern}
\usepackage{tablefootnote}

\usepackage{xcolor}

\title{ML-AutoResearch: Training Machine Learning Research Agents with Automatically Generated Environments}

\author{%
Ziyang Cai\thanks{Equal contribution. Work done during internship at Microsoft.} \\
Princeton University \\
\texttt{zc5794@princeton.edu} \\
\And
\text{Amir Saeidi}$^{*}$ \\
Arizona State University\\
\texttt{ssaeidi1@asu.edu} \\
\And
Harkirat Behl \\
Microsoft Research \\
\texttt{hbehl@microsoft.com} \\
}

\begin{document}

\maketitle\begin{abstract}

With the advent of AI agents, automated scientific discovery is becoming an increasingly plausible goal. However, training agents to autonomously execute the engineering-heavy labor of machine learning (ML) research requires massive, process-level supervision. Existing static benchmarks omit critical intermediate steps such as debugging and incremental reasoning, and manual data collection is prohibitively expensive. To overcome this data bottleneck, we introduce ML-AutoResearch (ML-AR), a scalable pipeline for automatically generating synthetic, end-to-end ML research tasks. Each task defines a complete research cycle, including problem specification, dataset selection, baseline implementation, and iterative improvement. To ensure realism and executability, tasks are grounded in real-world datasets and refined via an automated self-debugging procedure without requiring human supervision. We construct a large-scale dataset of teacher trajectories on these synthetic tasks to train student agents via supervised fine-tuning. We evaluate the resulting agents across 3 diverse ML research benchmarks. Our comprehensive experiments across 2 distinct model families and 3 model sizes demonstrate that training on ML-AR trajectories yields consistent and significant capability gains. Fine-tuning improves the Aggregation Under the Performance (AUP) by up to 9\% and substantially boosts overall pass rates, highlighting robust out-of-domain generalization.

\begin{figure}[h]
    \centering
    \includegraphics[width=0.9\linewidth]{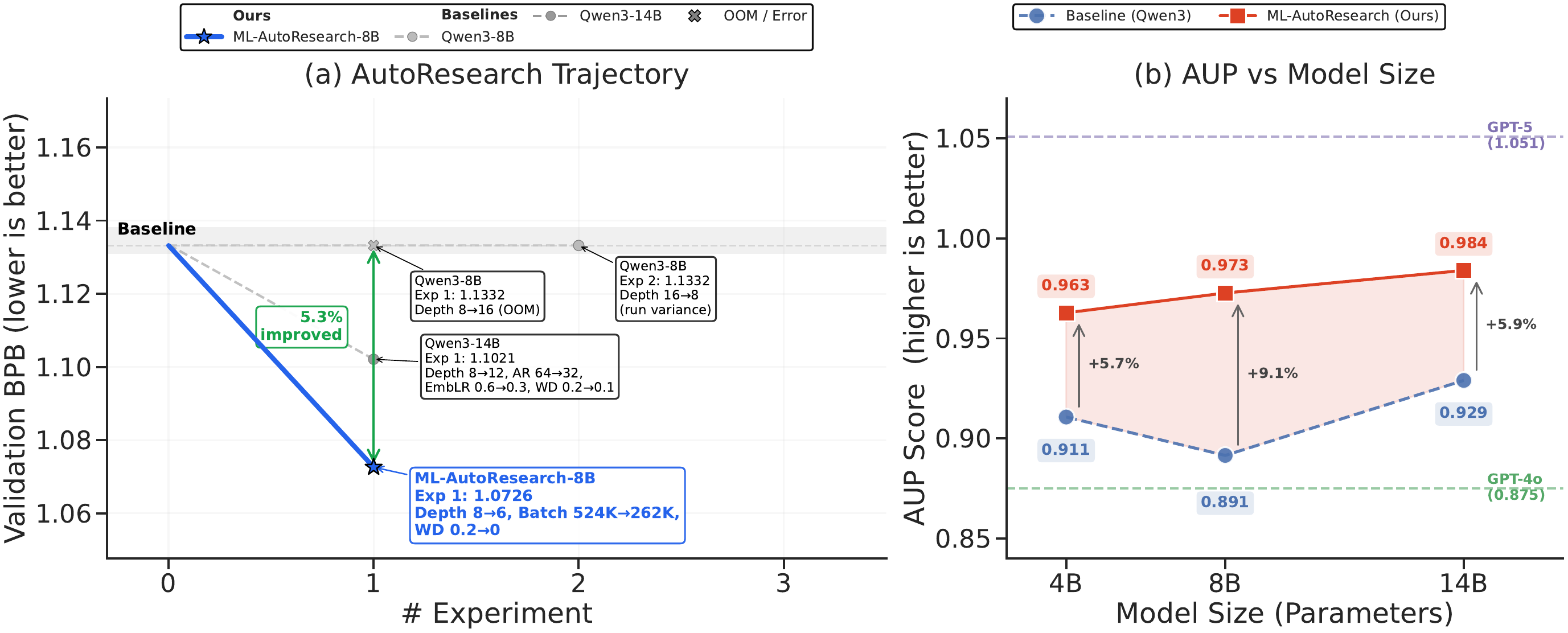}
    \caption{\textbf{Generalizability of ML-AutoResearch Models.} Performance comparison with the base instruction model across the MLGym and AutoResearch benchmarks.}
    \label{fig:dataset_comparision}
\end{figure}

\end{abstract}
\section{Introduction}

One of the longstanding goals of artificial intelligence is to enable autonomous scientific discovery. Recently, automating AI research has shifted from a long-term academic vision to an immediate industry focal point, driven by advanced coding capabilities that allow AI systems to reliably execute the engineering-heavy labor of machine learning development. Building autonomous systems to drive end-to-end AI R\&D is now a central pursuit for frontier laboratories, with academic prototypes like AI Scientist \cite{lu2024ai}, Co-Scientist \cite{gottweis2025towards}, and AlphaEvolve \cite{novikov2025alphaevolve} demonstrating early success. Although current large language models (LLMs) have internalized vast amounts of ML theory and programming knowledge \cite{brown2020language, touvron2023llama}, static comprehension alone is insufficient. Effective research requires the ability to iteratively act, evaluate, and refine decisions over extended horizons.

However, training capable ML agents to operate autonomously is fundamentally challenging. ML research is an open-ended, long-horizon endeavor requiring deep domain expertise, complex library-specific coding, and robust error recovery. Consequently, teaching an agent to navigate this space requires massive, diverse, and process-level supervision in the form of step-by-step reasoning and interaction traces. A central bottleneck in the field is the absence of such data. Gathering it manually is prohibitively expensive, and existing human-authored benchmarks expose only a handful of curated tasks, failing to provide the scale and diversity needed for stable agentic supervised fine-tuning.

Existing approaches to training coding and ML agents typically rely on static artifacts such as completed papers, code repositories, or static benchmarks \cite{yang2024swe, nathani2025mlgym}. These resources capture final outcomes but critically omit the intermediate steps essential for learning: failed experiments, debugging cycles, and incremental reasoning. While related efforts in agent training and tool use highlight the importance of interaction \cite{yao2022react, schick2023toolformer}, they lack scalable methods to generate diverse, process-level training data. As a result, agents trained on these signals struggle to translate static knowledge into effective, multi-step problem solving.

To address this limitation, we introduce \textbf{ML-AutoResearch (ML-AR)}, a scalable pipeline for generating synthetic ML research tasks together with rich agentic trajectories. Each generated task instantiates an end-to-end research cycle, including problem formulation, dataset selection, baseline implementation, and iterative improvement. The pipeline grounds tasks in real-world datasets through external verification and improves reliability via an automated self-debugging loop. Importantly, ML-AR is compatible with the task-agnostic SWE-agent framework \cite{yang2025swe}, enabling the generation of diverse research scenarios across domains with minimal manual effort.

We evaluate our approach on MLGym \cite{nathani2025mlgym}, MLAgentBench \cite{huang2023mlagentbench}, and AutoResearch, comprehensive benchmarks for ML agents comprising diverse tasks that require improving baseline implementations within a fixed interaction budget. In this setting, agents iteratively modify code, run experiments, and revise solutions based on execution feedback. From 1,000 sampled topics, ML-AR produces 555 verified task environments and approximately 34K teacher trajectories, which are filtered to 17K high-quality examples. The verified environments are not uniformly easy for GPT-5: it never improves the baseline on 195 tasks (35.1\%) and produces no valid solution on 72 tasks (13.0\%), indicating that the task pool spans its capability boundary. Across three evaluation seeds, fine-tuning Qwen3-4B, Qwen3-8B, and Qwen3-14B models \cite{yang2025qwen3} improves mean MLGym AUP by 5.7\%, 9.1\%, and 5.9\%, respectively. Mixing additional GPT-5.4 trajectories into the 14B training set further raises its AUP by 7.0\% over the baseline model. ML-AR also improves normalized MLAgentBench performance at every model size and transfers to AutoResearch. Finally, replacing GPT-5 with Qwen3-32B for trajectory generation still improves 8B and 14B students, showing that useful training trajectories are not specific to a proprietary teacher. Our primary contributions are summarized as follows:

% \begin{itemize}
%     \item We propose ML-AutoResearch, a scalable pipeline for generating synthetic, end-to-end ML research tasks grounded in real-world data without human supervision, yielding 555 verified environments from 1,000 sampled topics.
%     \item We construct a large dataset of agentic trajectories that captures experimentation, debugging, and feedback-driven refinement over long horizons.
%     \item We demonstrate aggregate gains across two model families, three model sizes, and three ML benchmarks, supported by multi-seed evaluation and train--test contamination analysis.
% \end{itemize}

\begin{itemize}
    \item We propose ML-AutoResearch, a scalable pipeline for generating synthetic, end-to-end ML research tasks grounded in real-world data without human supervision, including tasks that remain challenging for GPT-5.
    \item We construct a large dataset of agentic trajectories that capture iterative research behavior, including debugging and refinement over long horizons.
    \item We demonstrate consistent gains across two model families and three diverse ML benchmarks, including when Qwen3-32B replaces GPT-5 as the trajectory-generation teacher.
\end{itemize}

\begin{figure}[t!]
    \centering
    \includegraphics[width=1.0\textwidth]{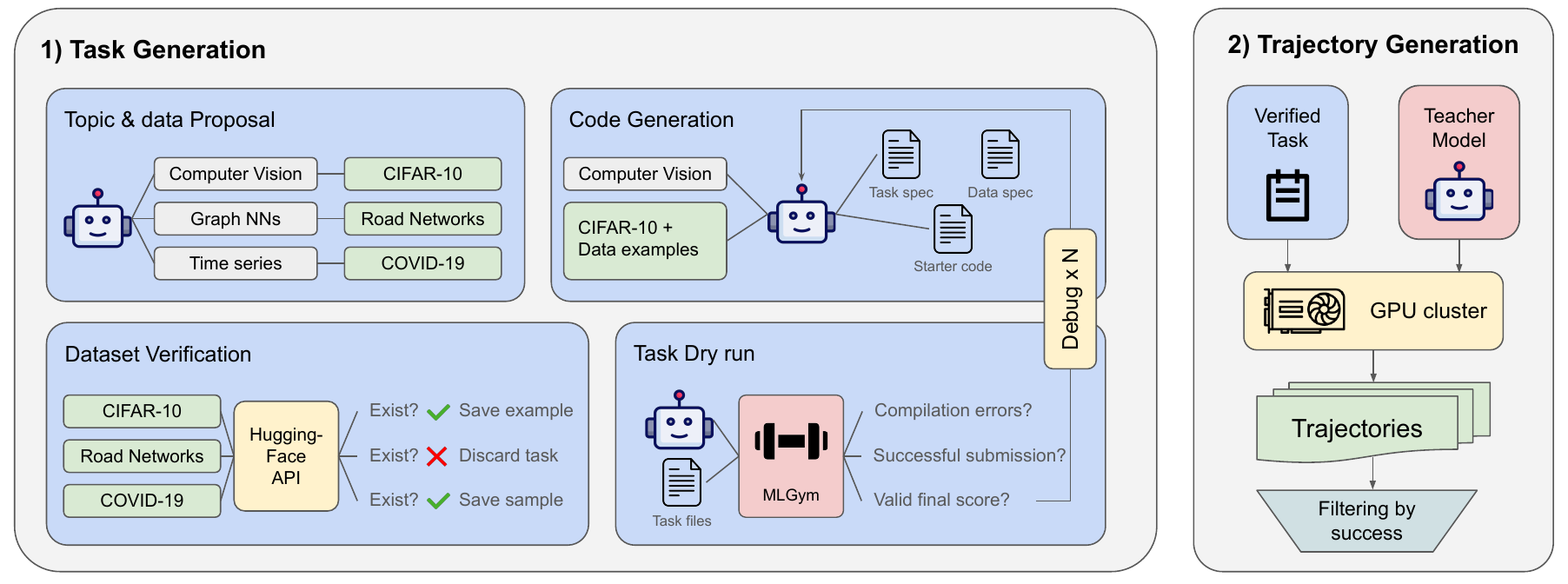}
    \caption{\textbf{Overview of the ML-AutoResearch framework}. \textbf{Stage 1:} Generate diverse research topic ideas. \textbf{Stage 2:} Verify each research topic by identifying a corresponding real-world dataset and discard topics for which no suitable dataset exists. \textbf{Stage 3:} Generate baseline code and evaluation pipelines for agents to solve the machine learning tasks. \textbf{Stage 4:} Execute tasks on a computational cluster, collect interaction trajectories, and select the best-performing trajectories.}
    \label{fig:myimage}
\end{figure}
\section{Related Work}

Recent work has increasingly explored the use of LLM-based agents to automate scientific research, ranging from ideation to end-to-end paper generation. Systems like AI Co-Scientist \cite{gottweis2025towards} and Bit-Flip \cite{o2025sparks} focus heavily on generating novel hypotheses, while frameworks such as The AI Scientist \cite{lu2024ai} attempt full hypothesis-to-paper research loops. However, downstream studies reveal a pronounced ideation-execution gap, where models propose novel ideas but struggle to reliably implement them \cite{si2025ideation, siegel2024core}. To measure these execution capabilities, several benchmarks have been introduced, including MLE-Bench for Kaggle-style engineering \cite{chan2024mle}, PaperBench for paper replication \cite{starace2025paperbench}, and long-horizon environments like MLAgentBench, MLGym, and MLRC-Bench \cite{huang2023mlagentbench, nathani2025mlgym, zhang2025mlrc, chen2025mlr}. These evaluations consistently show that while agents can navigate established pipelines, they struggle with robust planning and genuinely novel method discovery over long horizons.

To address these execution shortcomings, there is a growing need for scalable methods to generate process-level training data. Because human-authored benchmarks are too small for stable agentic fine-tuning, recent efforts have turned to synthetic data generation. In software engineering, for example, SWE-Smith scales task generation by synthesizing test-breaking instances across Python codebases to improve downstream agent performance \cite{yang2025swe}. Similarly, automated loops have been used to generate targeted optimization tasks, such as NanoGPT speedruns \cite{zhao2025automated}. Our work, ML-AutoResearch, builds directly upon these paradigms. By algorithmically synthesizing and verifying end-to-end ML research tasks, we provide the massive, process-level supervision required to bridge the execution gap and improve autonomous discovery on rigorous benchmarks like MLGym and MLAgentBench.

\section{ML-AutoResearch Framework}
\label{sec:method}

% \HB{What is ML Agent, why is it useful}

% \HB{Why is this problem difficult}

% \HB{Inspiration/Motivation of method. How did you come up with.}

Machine learning (ML) agents hold the potential to accelerate scientific discovery by autonomously conducting ML workflows through code editors and command-line interfaces. However, training these agents is fundamentally challenging. Navigating open-ended, long-horizon ML research requires deep domain expertise and robust error recovery, demanding massive, diverse, process-level supervision for supervised fine-tuning (SFT). A central bottleneck in the field is the absence of such data: manual collection is prohibitively expensive, and existing human-authored benchmarks expose only a handful of curated tasks, failing to provide the necessary scale.

Inspired by recent successes in synthetic data generation, we hypothesize that the vast repositories of open-source datasets can be transformed into an infinite curriculum of structured ML tasks. Our core insight is to utilize a highly capable ``teacher'' LLM as an automated environment architect to dynamically synthesize high-quality, executable ML research environments at scale without human annotation. To this end, we introduce ML-AutoResearch (ML-AR), an end-to-end pipeline organized into three phases: (1) environment synthesis, which proposes tasks grounded in real-world datasets, (2) environment verification, which filters out ill-posed tasks via a teacher-driven self-debugging loop, and (3) trajectory generation and filtering, which curates large-scale agent rollouts for downstream training. We first formalize the problem setting and then describe each phase in detail. An overview is given in Figure~\ref{fig:myimage}.

\begin{figure}
    \centering
    \includegraphics[width=1\linewidth]{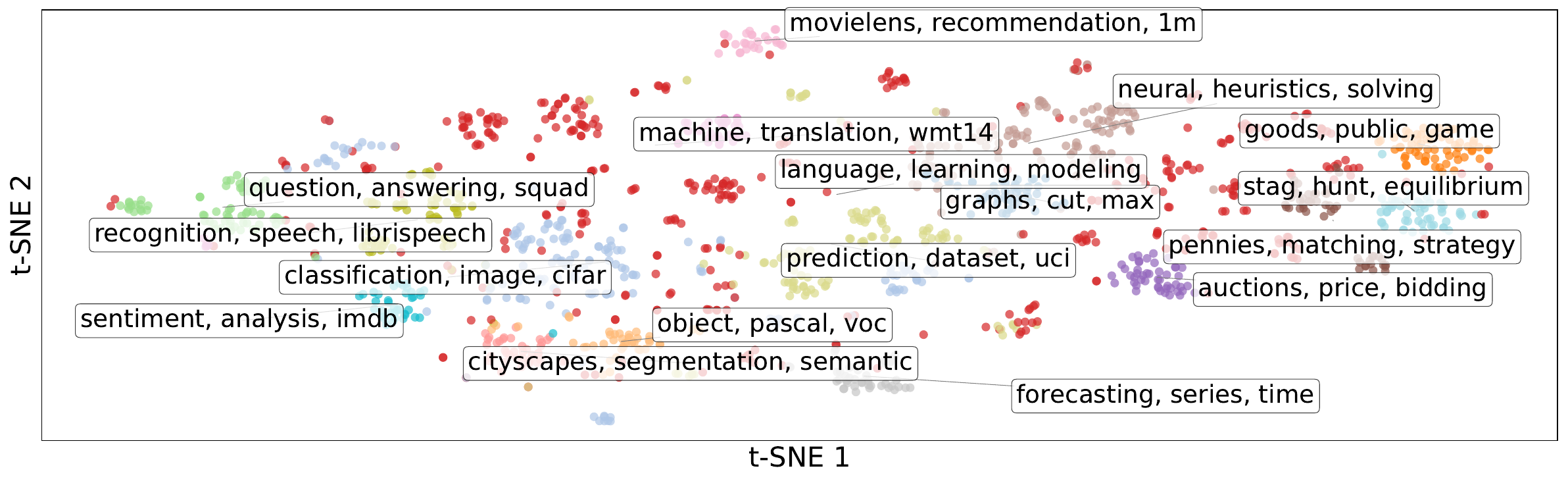}
    \caption{\textbf{Demonstration of Research Topic Diversity.} Overview of the diversity of topics generated during topic sampling.}
    \label{fig:topic_diversity}
\end{figure}

\subsection{Problem Statement}
\label{sec:problem_statement}

We consider an agent that interacts with an ML research environment in the SWE-agent interaction format adopted by MLGym. A task is a tuple $T = (q, d, \mathcal{C}, \mathcal{V})$, where $q$ is a natural-language specification of a machine learning problem, $d$ is a dataset descriptor, $\mathcal{C}$ is a collection of starter and helper code files providing a baseline implementation, and $\mathcal{V}$ is an evaluation procedure that maps any candidate submission to a scalar score $s\in\mathbb{R}$. Given $T$, an agent $\pi$ interacts with an execution environment $\mathcal{E}_T$ for at most $H$ rounds; at round $t$ it observes a context $o_t$ (file contents, previous tool outputs, evaluator feedback) and produces a reasoning trace and an action $(h_t, a_t) \sim \pi(\cdot \mid o_{\le t}, a_{<t})$, where $a_t$ may inspect or edit files, run shell commands inside a sandboxed virtual environment, or submit a candidate solution; multiple submissions are permitted within the budget $H$. The interaction yields a trajectory $\tau = (o_0, h_1, a_1, o_1, \ldots, h_{H'}, a_{H'}, o_{H'})$ with $H' \le H$, whose performance on $T$ is summarized by the best score $s^\star(\tau, T) = \max_{a_t \in \tau:\,\text{submission}} \mathcal{V}(a_t)$. We assume access to a strong teacher policy $\pi_{\mathrm{T}}$ (in our case GPT-5) and aim to train a student policy $\pi_\theta$ via maximum-likelihood imitation of teacher trajectories,
\begin{equation}
    \min_{\theta}\; \mathbb{E}_{T\sim p(T)}\,\mathbb{E}_{\tau\sim p_{\pi_{\mathrm{T}}}(\tau\mid T)}
    \!\left[\, -\!\!\sum_{t=1}^{H'}\log \pi_\theta\bigl(h_t,\,a_t \mid o_{\le t},\, a_{<t}\bigr)\,\right],
    \label{eq:sft}
\end{equation}
% restricted to trajectories that pass the filter described in Section~\ref{sec:phase3}. Solving Eq.~\ref{eq:sft} requires a sufficiently large and diverse task distribution $p(T)$, but human-curated benchmarks define $p(T)$ on at most a few dozen tasks, which is far too narrow for stable agentic SFT. The objective of ML-AutoResearch is therefore to construct $p(T)$ algorithmically: starting only from the teacher $\pi_{\mathrm{T}}$ and the open HuggingFace dataset hub $\mathcal{H}$, we induce a distribution $\widehat{p}(T) = p(T \mid \pi_{\mathrm{T}}, \mathcal{H})$ over executable, dataset-grounded ML research tasks and use it as a surrogate for $p(T)$ in Eq.~\ref{eq:sft}. The remainder of this section describes how $\widehat{p}(T)$ is sampled, verified, and used to generate trajectories.

Restricted to trajectories that pass the filter described in Section~\ref{sec:phase3}, optimizing Eq.~\ref{eq:sft} requires a large and diverse task distribution $p(T)$. However, human-curated benchmarks cover only a limited set of tasks, making them insufficient for stable agentic SFT. To address this, ML-AutoResearch constructs $p(T)$ algorithmically by inducing a distribution $\widehat{p}(T) = p(T \mid \pi_{\mathrm{T}}, \mathcal{H})$ over executable, dataset-grounded ML research tasks using the teacher $\pi_{\mathrm{T}}$ and the HuggingFace dataset hub $\mathcal{H}$. This induced distribution serves as a surrogate for $p(T)$ in Eq.~\ref{eq:sft}, and is used to sample, verify, and generate trajectories.

\begin{algorithm}[htbp]
\caption{Self-debugging verification loop for a candidate task $T_i$.}
\label{alg:verify-text}
\begin{algorithmic}[1]
\State \textbf{Input:} candidate $T_i$, teacher $\pi_{\mathrm{T}}$, max iterations $K$, debug probability $p_{\mathrm{debug}}$
\For{$j = 1, \ldots, K$}
    \State Execute $\pi_{\mathrm{T}}$ on $T_i$ in $\mathcal{E}_{T_i}$, collect error log $\varepsilon_j$
    \If{$\varepsilon_j = \varnothing$}
        \State \Return $T_i$ \Comment{verified}
    \EndIf
    \State Sample $u \sim \mathrm{Uniform}(0,1)$
    \State \textbf{Regeneration step}
    \If{$u < p_{\mathrm{debug}}$}
        \State Regenerate $\mathcal{C}_i \sim \pi_{\mathrm{T}}(\mathcal{C}\mid q_i, d_i, \varepsilon_j)$ \Comment{repair}
    \Else
        \State Regenerate $\mathcal{C}_i \sim \pi_{\mathrm{T}}(\mathcal{C}\mid q_i, d_i)$ \Comment{restart}
    \EndIf
\EndFor
\State \Return discard
\end{algorithmic}
\end{algorithm}

\subsection{Phase 1: Environment Synthesis}
\label{sec:phase1}
The first phase samples candidate tasks from $\widehat{p}(T)$. To encourage breadth without sacrificing executability, we factor task generation into three conditional steps that are each implemented as a structured prompt to the teacher $\pi_{\mathrm{T}}$.
% The prompt templates used in our pipeline are provided in Appendix~\ref{sec:appendix-prompts}.

\paragraph{Stage 1: Topic sampling.} We first draw a set of $N$ machine learning topics $\mathcal{Z} = \{z_1, \ldots, z_N\} \sim \pi_{\mathrm{T}}(z)$, where each $z_i$ is a short natural-language descriptor of a research area (e.g.\ contrastive sentence embedding, multi-agent matrix games, tabular anomaly detection). To promote diversity we sample topics in batches and include previously sampled topics in the prompt as exclusions (See Figure \ref{fig:topic_diversity}). Topics serve as the only source of stochasticity in the pipeline; everything downstream is conditioned on the chosen $z_i$.

\paragraph{Stage 2: Task and dataset proposal.} For each topic $z_i$, the teacher proposes a task description $q_i$ together with a candidate dataset name $\tilde d_i$, drawn jointly as $(q_i, \tilde d_i) \sim \pi_{\mathrm{T}}(\cdot \mid z_i)$. We then ground $\tilde d_i$ in a real-world artifact by issuing a query to the HuggingFace search API and selecting the closest match $d_i \in \mathcal{H}$. If the match score falls below a fixed similarity threshold, the task is discarded; if it succeeds, we enrich $d_i$ with a small set of example rows fetched from HuggingFace, which allows the teacher to write code that respects the actual schema of the data. Topics that do not require a dataset (e.g.\ game-theoretic environments) bypass this grounding step and proceed with $d_i = \varnothing$.

\paragraph{Stage 3: Configuration and starter-code generation.} Conditioned on the grounded pair $(q_i, d_i)$, the teacher synthesizes the remaining components of the task as $(\mathcal{C}_i, \mathcal{V}_i, \kappa_i) \sim \pi_{\mathrm{T}}(\cdot \mid q_i, d_i)$, where $\mathcal{C}_i$ is the starter and helper code, $\mathcal{V}_i$ is an evaluation script, and $\kappa_i$ is a configuration file describing how MLGym should mount the task (entry points, time limits, metric direction). The output of Phase~1 is a candidate task $T_i = (q_i, d_i, \mathcal{C}_i, \mathcal{V}_i)$ together with $\kappa_i$, which is passed to Phase~2 for verification.

\subsection{Phase 2: Environment Verification}
\label{sec:phase2}
Because every step of Phase~1 may introduce errors, naively running an agent on a candidate $T_i$ is unreliable: code may fail to compile, datasets may not match the assumed schema, or the evaluator may be ill-defined. We therefore subject each candidate to a teacher-driven verification loop that explicitly tests end-to-end executability before the task is admitted to $\widehat{p}(T)$.

Concretely, we plug $T_i$ into MLGym and run a single rollout of the teacher $\pi_{\mathrm{T}}$ for at most $H$ rounds. Let $E_i \in \{0, 1\}$ indicate whether the rollout produced an exception or a non-finite evaluator score. If $E_i = 0$, we accept $T_i$ together with the resulting baseline trajectory and admit it to the verified task pool $\widehat{\mathcal{T}}$. Phase~2 tests only whether the environment executes end to end and returns a finite score; it does not require the teacher to outperform the baseline. Baseline improvement is considered only during Phase~3 trajectory filtering. If $E_i = 1$, we initiate a self-debugging loop, summarized in Algorithm~\ref{alg:verify-text}.

% \begin{algorithm}[htbp]
% \caption{Self-debugging verification loop for a candidate task $T_i$.}
% \label{alg:verify-text}
% \begin{algorithmic}[1]
% \State \textbf{Input:} candidate $T_i$, teacher $\pi_{\mathrm{T}}$, max iterations $K$, debug probability $p_{\mathrm{debug}}$
% \For{$j = 1, \ldots, K$}
%     \State Execute $\pi_{\mathrm{T}}$ on $T_i$ in $\mathcal{E}_{T_i}$, collect error log $\varepsilon_j$
%     \If{$\varepsilon_j = \varnothing$}
%         \State \Return $T_i$ \Comment{verified}
%     \EndIf
%     \State Sample $u \sim \mathrm{Uniform}(0,1)$
%     \If{$u < p_{\mathrm{debug}}$}
%         \State Regenerate $\mathcal{C}_i \sim \pi_{\mathrm{T}}(\mathcal{C}\mid q_i, d_i, \varepsilon_j)$ \Comment{repair}
%     \Else
%         \State Regenerate $\mathcal{C}_i \sim \pi_{\mathrm{T}}(\mathcal{C}\mid q_i, d_i)$ \Comment{restart}
%     \EndIf
% \EndFor
% \State \Return discard
% \end{algorithmic}
% \end{algorithm}

The probability $p_{\mathrm{debug}}$ trades off two failure modes: pure repair ($p_{\mathrm{debug}}=1$) tends to lock onto a flawed code skeleton, whereas pure restart ($p_{\mathrm{debug}}=0$) discards potentially useful partial progress. In practice we find that intermediate values strike a good balance and noticeably increase the yield of verified tasks per topic. Critically, the entire verification phase requires no human input and parallelizes trivially across tasks, so the size of $\widehat{\mathcal{T}}$ scales with available compute rather than with annotation budget.

\paragraph{Verification yield and task difficulty.}
Of 1,000 sampled topics, 914 passed dataset validation and entered Phase~2; 86 were removed because no valid dataset was found. Of these candidates, 555 passed environment verification (60.7\%), while 359 were discarded because repeated debugging did not produce an executable environment or valid configuration. Table~\ref{tab:task_difficulty_distribution} characterizes the verified tasks using up to 256 GPT-5 trajectories per task. GPT-5 improved the baseline at least once on 360/555 tasks (64.9\%) and produced a valid solution on 483/555 tasks (87.0\%). The substantial mass in both intermittent-success and unsuccessful categories shows that verification retains tasks spanning the teacher's capability boundary rather than only tasks it reliably improves.

\begin{table}[t]
    \centering
    \small
    \begin{tabular}{lrr}
        \toprule
        GPT-5 outcome across rollouts & Tasks & Share \\
        \midrule
        Consistently improves baseline & 175 & 31.5\% \\
        Improves in some rollouts      & 185 & 33.3\% \\
        Valid, but never improves      & 123 & 22.2\% \\
        No valid solution              & 72  & 13.0\% \\
        \midrule
        Total verified tasks           & 555 & 100.0\% \\
        \bottomrule
    \end{tabular}
    \caption{\textbf{Difficulty distribution of verified tasks.} Outcomes are measured across up to 256 GPT-5 trajectories per task. The distribution spans tasks that GPT-5 improves reliably, improves only intermittently, or cannot improve or solve.}
    \label{tab:task_difficulty_distribution}
\end{table}

\subsection{Phase 3: Trajectory Generation and Filtering}
\label{sec:phase3}
Given the verified task pool $\widehat{\mathcal{T}}$, the final phase produces the trajectory dataset that drives the SFT objective in Eq.~\ref{eq:sft}.

\paragraph{Large-scale sampling.} For each verified task $T_i\in\widehat{\mathcal{T}}$ we draw $K$ independent teacher trajectories $\tau_i^{(k)} \sim p_{\pi_{\mathrm{T}}}(\tau \mid T_i)$ for $k = 1, \ldots, K$, by running $\pi_{\mathrm{T}}$ in $\mathcal{E}_{T_i}$ on an HPC cluster, with one GPU allocated per concurrent rollout. We target $K = 256$ per task; however, because the pipeline is unsupervised, individual tasks can fail intermittently due to evaluator timeouts, file-system contention, or container instabilities, so the realized number of successful rollouts $K_i \le K$ varies across tasks.

% \paragraph{Trajectory filtering.} After sampling the raw rollouts, we must identify and extract the highest-quality trajectories to construct the supervised fine-tuning dataset. At this stage, we evaluate two distinct performance-based selection strategies. The first strategy, Z-score sampling, calculates the standard score of the primary metric for each trajectory. Specifically, we compute $z_i^{(k)} = (s_i^{(k)} - \mu_i) / \sigma_i$, where $s_i^{(k)}$ is the primary score of the trajectory, and $\mu_i$ and $\sigma_i$ are the empirical mean and standard deviation of valid submissions for task $T_i$. We then discard trajectories falling below a predefined threshold $z_{\mathrm{th}}$. The second strategy, successful experience selection, strictly retains only those trajectories that achieve a metric score superior to the baseline implementation. Finally, after applying the chosen selection strategy, we filter the dataset to include only trajectories containing fewer than $32{,}768$ tokens. This strict length constraint ensures that the model trains exclusively on tasks that were completely solved within the allocated context window, preserving the full reasoning and execution history without relying on truncation. The finalized dataset $\mathcal{D}_{\mathrm{SFT}}$ comprises all pairs $(T_i, \tau_i^{(k)})$ that pass these criteria, forming the empirical distribution used to optimize the SFT objective in Eq.~\ref{eq:sft}. The statistical details of the training set are shown in Figure \ref{fig:dataset_stats}.

\begin{figure}[t!]
    \centering
    \includegraphics[width=1\linewidth]{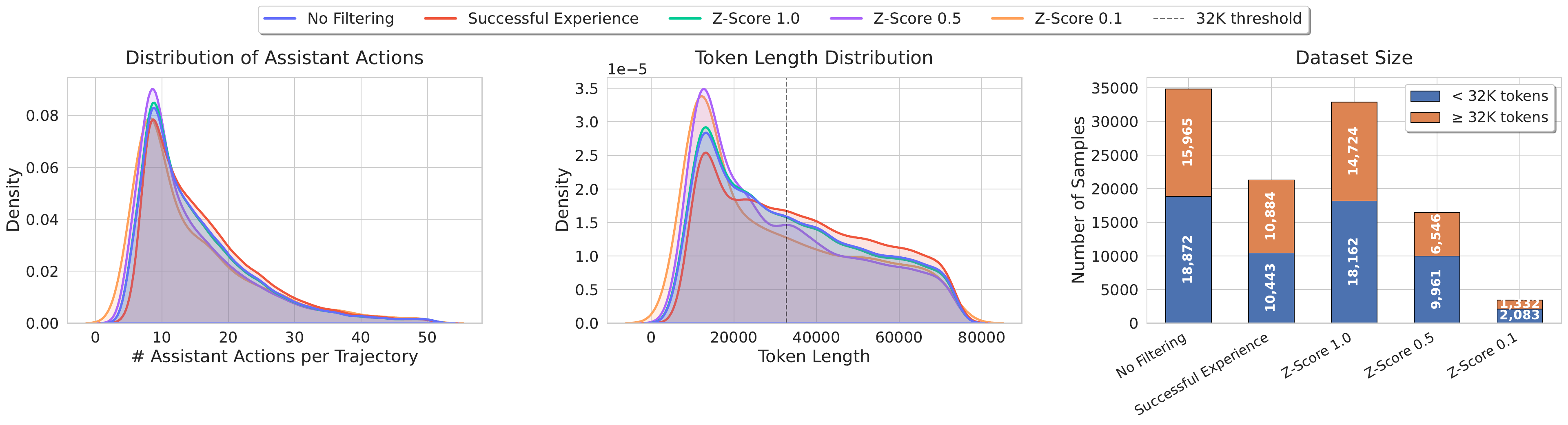}
    \caption{\textbf{Dataset Statistics.} Overview of the training set constructed by the ML-AutoResearch framework across different settings.}
    \label{fig:dataset_stats}
\end{figure}

\paragraph{Trajectory filtering.} After sampling raw rollouts, we select high-quality trajectories to construct the supervised fine-tuning dataset using two performance-based strategies. First, Z-score sampling computes $z_i^{(k)} = (s_i^{(k)} - \mu_i) / \sigma_i$ and discards trajectories below a threshold $z_{\mathrm{th}}$. Second, successful experience selection retains only trajectories that outperform the baseline. We further filter trajectories to those under $32{,}768$ tokens to ensure complete reasoning within the context window. The final dataset $\mathcal{D}_{\mathrm{SFT}}$ consists of all $(T_i, \tau_i^{(k)})$ pairs that satisfy these criteria. Statistical details are shown in Figure \ref{fig:dataset_stats}.

\section{Experiments}
% \paragraph{Setup}
In this section, we evaluate the ML-AutoResearch (ML-AR) framework across three diverse benchmarks to demonstrate its generalizability and scaling properties across different sizes and families of Qwen models.
\paragraph{Models and Environments.} 
We utilize Qwen3-4B, Qwen3-8B, Qwen3-14B, and Qwen2.5-3B-Instruct as base models, fine-tuning them on the synthetic trajectory dataset generated by our ML-AutoResearch pipeline. We benchmark these fine-tuned student models against their respective un-tuned base models, as well as against frontier proprietary models, specifically GPT-4o and GPT-5.

\paragraph{Benchmarks.} 
We evaluate the models on three comprehensive benchmarks: MLGym, AutoResearch, and MLAgentBench, which collectively span a wide spectrum of machine learning research tasks. Specifically, \textbf{MLGym} provides a broad evaluation suite with a strong emphasis on reinforcement learning environments, while also covering general ML engineering tasks. \textbf{AutoResearch} focuses on iterative research and optimization workflows, tasking the agent to tune a Small Language Model (SLM) to achieve a low validation bits-per-byte (\texttt{val\_bpb}) metric. Finally, \textbf{MLAgentBench} offers a complementary suite of diverse, open-ended machine learning design and debugging tasks. For further details regarding the benchmark configurations and evaluation metrics, refer to Appendix~\ref{sec:appendix_benchmarks_details}.

% \begin{table}[t]
%     \centering
%     \begin{tabular}{lccc}
%         \toprule
%         Model size & Qwen3-Instruct & ML-AR & Absolute gain \\
%         \midrule
%         4B  & $0.000 \pm 0.000$ & $0.101 \pm 0.100$ & $+0.101$ \\
%         8B  & $0.044 \pm 0.077$ & $0.215 \pm 0.212$ & $+0.171$ \\
%         14B & $0.126 \pm 0.127$ & $0.420 \pm 0.193$ & $+0.294$ \\
%         \bottomrule
%     \end{tabular}
%     \caption{\textbf{Three-seed MLAgentBench evaluation.} Mean normalized score and standard deviation across evaluation seeds 42, 123, and 456. Differences are absolute normalized-score gains.}
%     \label{tab:mlagentbench_multiseed}
% \end{table}

% \begin{table}[t]
%     \centering
%     \small
%     \begin{NiceTabular}{lccc}
%         \toprule
%         Model & Qwen3 & ML-AR & Absolute gain \\
%         \midrule
%         4B  & $0.000_{\scriptscriptstyle \pm 0.000}$
%             & $0.101_{\scriptscriptstyle \pm 0.100}$
%             & \textbf{$+0.101$} \\
%         8B  & $0.044_{\scriptscriptstyle \pm 0.077}$
%             & $0.215_{\scriptscriptstyle \pm 0.212}$
%             & \textbf{$+0.171$} \\
%         14B & $0.126_{\scriptscriptstyle \pm 0.127}$
%             & $0.420_{\scriptscriptstyle \pm 0.193}$
%             & \textbf{$+0.294$} \\
%         \bottomrule
%     \end{NiceTabular}
%     \vspace{3mm}
%     \caption{\textbf{Three-seed MLAgentBench evaluation.} Mean normalized score and standard deviation across evaluation seeds 42, 123, and 456. Differences are absolute normalized-score gains.}
%     \label{tab:mlagentbench_multiseed}
% \end{table}

\begin{table}[t]
    \centering

    % -------- Left table --------
    \begin{minipage}[t]{0.49\columnwidth}
        \centering
        \scriptsize
        \resizebox{\linewidth}{!}{%
        \begin{NiceTabular}{lccc}
            \toprule
            Model & Qwen3 & ML-AR & Rel. gain \\
            \midrule
            4B  & $0.9107_{\scriptscriptstyle \pm 0.0099}$ 
                & $0.9628_{\scriptscriptstyle \pm 0.0237}$ 
                & \textbf{$+5.7\%$} \\
            8B  & $0.8915_{\scriptscriptstyle \pm 0.0371}$ 
                & $0.9727_{\scriptscriptstyle \pm 0.0656}$ 
                & \textbf{$+9.1\%$} \\
            14B & $0.9290_{\scriptscriptstyle \pm 0.0204}$ 
                & $0.9840_{\scriptscriptstyle \pm 0.0364}$ 
                & \textbf{$+5.9\%$} \\
            \bottomrule
        \end{NiceTabular}%
        }
        \vspace{3mm}
        \captionof{table}{\textbf{Three-seed MLGym evaluation.} Mean AUP and standard deviation across evaluation seeds 42, 123, and 456.}
        \label{tab:mlgym_multiseed}
    \end{minipage}
    \hfill
    % -------- Right table --------
    \begin{minipage}[t]{0.49\columnwidth}
        \centering
        \scriptsize
        \resizebox{\linewidth}{!}{%
        \begin{NiceTabular}{lccc}
            \toprule
            Model & Qwen3 & ML-AR & Abs. gain \\
            \midrule
            4B  & $0.000_{\scriptscriptstyle \pm 0.000}$
                & $0.101_{\scriptscriptstyle \pm 0.100}$
                & \textbf{$+0.101$} \\
            8B  & $0.044_{\scriptscriptstyle \pm 0.077}$
                & $0.215_{\scriptscriptstyle \pm 0.212}$
                & \textbf{$+0.171$} \\
            14B & $0.126_{\scriptscriptstyle \pm 0.127}$
                & $0.420_{\scriptscriptstyle \pm 0.193}$
                & \textbf{$+0.294$} \\
            \bottomrule
        \end{NiceTabular}%
        }
        \vspace{3mm}
        \captionof{table}{\textbf{Three-seed MLAgentBench evaluation.} Mean normalized score and standard deviation across seeds 42, 123, and 456.}
        \label{tab:mlagentbench_multiseed}
    \end{minipage}

\end{table}

\subsection{Main Results}

\paragraph{ML-AutoResearch improves aggregate MLGym performance across model sizes.}
We reevaluate each base and ML-AR checkpoint using evaluation seeds 42, 123, and 456. As shown in Table~\ref{tab:mlgym_multiseed}, mean AUP improves by 5.7\%, 9.1\%, and 5.9\% at 4B, 8B, and 14B, respectively. The gains persist across all three model sizes, although performance remains heterogeneous across individual tasks. Independent training-seed experiments further yield mean AUP gains of 4.4\% at 8B and 4.0\% at 14B (Appendix~\ref{sec:appendix-training-seeds}).

% \begin{table}[t]
%     \centering
%     \begin{tabular}{lcc}
%         \toprule
%         Model (seed 42) & Best-run AUP & Gain over base \\
%         \midrule
%         Qwen3-14B & 1.0555 & --- \\
%         Original ML-AR-14B & 1.0740 & $+1.8\%$ \\
%         ML-AR-14B-Mixed & \textbf{1.1297} & $\mathbf{+7.0\%}$ \\
%         \bottomrule
%     \end{tabular}
%     \caption{\textbf{Effect of additional data at 14B.} Seed-42 best-run AUP. ML-AR-14B-Mixed combines the original GPT-5 trajectories with additional GPT-5.4 trajectories and improves by 5.2\% over the original ML-AR-14B.}
%     \label{tab:additional_data_14b}
% \end{table}

\begin{table}[h]
    \centering
    \small
    \begin{NiceTabular}{lcc}
        \toprule
        Model & Best-run AUP & Gain over base \\
        \midrule
        Qwen3-14B & 1.0555 & --- \\
        Original ML-AR-14B & 1.0740 & $+1.8\%$ \\
        ML-AR-14B-Mixed & \textbf{1.1297} & \textbf{$+7.0\%$} \\
        \bottomrule
    \end{NiceTabular}
    \vspace{2mm}
    \caption{\textbf{Effect of additional data at 14B.} Seed-42 best-run AUP. ML-AR-14B-Mixed combines the original GPT-5 trajectories with additional GPT-5.4 trajectories and improves by 5.2\% over the original ML-AR-14B.}
    \label{tab:additional_data_14b}
\end{table}

\paragraph{Additional trajectory data improves the 14B model.}
For the final 14B model, we mix additional GPT-5.4 trajectories with the original GPT-5 trajectory dataset. Under the seed-42 best-run protocol, this ML-AR-14B-Mixed model reaches an AUP of 1.1297, improving by 7.0\% over Qwen3-14B and by 5.2\% over the original ML-AR-14B (Table~\ref{tab:additional_data_14b}). We therefore use ML-AR-14B-Mixed as the primary 14B model in subsequent results and retain the original ML-AR-14B only for this data-scaling comparison. These best-run values are separate from the three-seed means in Table~\ref{tab:mlgym_multiseed}.

% \begin{table*}[t]
%     \centering
%     \scriptsize
%     \begin{tabular}{llccccc}
%         \toprule
%         Size & Model & Seed 42 & Seed 123 & Seed 456 & AUP mean $\pm$ std & Reletive gain \\
%         \midrule
%         8B  & Qwen3-Instruct & 1.5755 & 0.6331 & 0.8523 & $1.0203 \pm 0.4027$ & -- \\
%         8B  & ML-AR          & 1.7464 & 0.5522 & 0.8529 & $1.0505 \pm 0.5072$ & $+3.0\%$ \\
%         14B & Qwen3-Instruct & 1.7129 & 0.6218 & 0.9121 & $1.0823 \pm 0.4614$ & -- \\
%         14B & ML-AR          & 1.7403 & 0.6353 & 0.9764 & $1.1173 \pm 0.4620$ & $+3.2\%$ \\
%         \bottomrule
%     \end{tabular}
%     \caption{\textbf{Open-weight trajectory teacher.} MLGym AUP by evaluation seed and aggregate performance after fine-tuning on trajectories generated by Qwen3-32B. The verified task environments are held fixed.}
%     \label{tab:open_weight_teacher}
% \end{table*}

\begin{table*}[t]
    \centering
    \scriptsize
    \begin{NiceTabular}{llccccc}
        \toprule
        Size & Model & \multicolumn{3}{c}{Per-seed AUP} & Mean AUP & Rel. gain \\
        \cmidrule(lr){3-5}
             &       & 42 & 123 & 456 & & \\
        \midrule

        8B  & Qwen3
            & 1.5755 & 0.6331 & 0.8523
            & $1.0203_{\scriptscriptstyle \pm 0.4027}$
            & -- \\

        8B  & ML-AR
            & 1.7464 & 0.5522 & 0.8529
            & $1.0505_{\scriptscriptstyle \pm 0.5072}$
            & \textbf{$+3.0\%$} \\

        14B & Qwen3
            & 1.7129 & 0.6218 & 0.9121
            & $1.0823_{\scriptscriptstyle \pm 0.4614}$
            & -- \\

        14B & ML-AR
            & 1.7403 & 0.6353 & 0.9764
            & $1.1173_{\scriptscriptstyle \pm 0.4620}$
            & \textbf{$+3.2\%$} \\

        \bottomrule
    \end{NiceTabular}
    \caption{\textbf{Open-weight trajectory teacher.} MLGym AUP by evaluation seed and aggregate performance after fine-tuning on trajectories generated by Qwen3-32B. The verified task environments are held fixed.}
    \label{tab:open_weight_teacher}
\end{table*}

\paragraph{Open-weight trajectory generation also improves student models.}
To test whether the training gains depend specifically on GPT-5-generated behavior, we use Qwen3-32B to generate trajectories on the same 555 verified environments. Qwen3-32B generates 6K trajectories, of which 2.1K remain after filtering. Across three evaluation seeds, models trained on these trajectories improve mean MLGym AUP by 3.0\% at 8B and 3.2\% at 14B (Table~\ref{tab:open_weight_teacher}). This experiment replaces the trajectory-generation teacher only; task synthesis and verification still use GPT-5. Full setup and task-level results appear in Appendix~\ref{sec:appendix-open-weight-teacher}.

\paragraph{ML-AutoResearch transfers to MLAgentBench.}
Across evaluation seeds 42, 123, and 456, ML-AR improves mean normalized performance at every model size (Table~\ref{tab:mlagentbench_multiseed}). The absolute gains are 0.101, 0.171, and 0.294 at 4B, 8B, and 14B, respectively, with the largest transfer gain at 14B. These results span different codebases, objectives, and evaluators from the synthetic training environments.

\paragraph{ML-AutoResearch-tuned models reliably discover superior solutions on the first attempt.} 
To further demonstrate the out-of-domain generalizability of the model, we evaluated the ML-AR tuned agents on the AutoResearch benchmark. In this setting, the agent is provided with the training script of a small language model, alongside configurations and resource constraints, and must autonomously discover optimal hyperparameters to surpass a baseline metric. Our results demonstrate that the ML-AutoResearch model, upon analyzing the training code and initial observations, consistently proposes superior hyperparameter configurations on its first attempt. In particular, the tuned models achieve a 5\% improvement on the \texttt{val\_bpb} metric compared to baselines (See Figure~\ref{fig:dataset_comparision}). This zero-shot capability to parse execution contexts and immediately produce performance-enhancing modifications highlights the robust, creative reasoning unlocked by our framework.

\paragraph{Training tasks have limited overlap with evaluation tasks.}
We compare synthetic task names and descriptions against MLGym and MLAgentBench and find no shared 8-grams. We then encode each task with \texttt{nvidia/llama-embed-nemotron-8b} and compute pairwise cosine similarities. Average train--test similarity is below 0.30; only 6 of 555 synthetic tasks (1.1\%) have maximum similarity above 0.60, all with MLGym, and none exceed 0.70 (Table~\ref{tab:contamination}). Trajectories from the six tasks account for less than 0.9\% of the training set. Although familiarity with the MLGym interaction format may still contribute to performance, transfer to MLAgentBench and the structurally distinct AutoResearch workflow suggests that interface alignment is not the sole explanation for the gains.

\begin{figure}[t]
    \centering
    \includegraphics[width=1\linewidth]{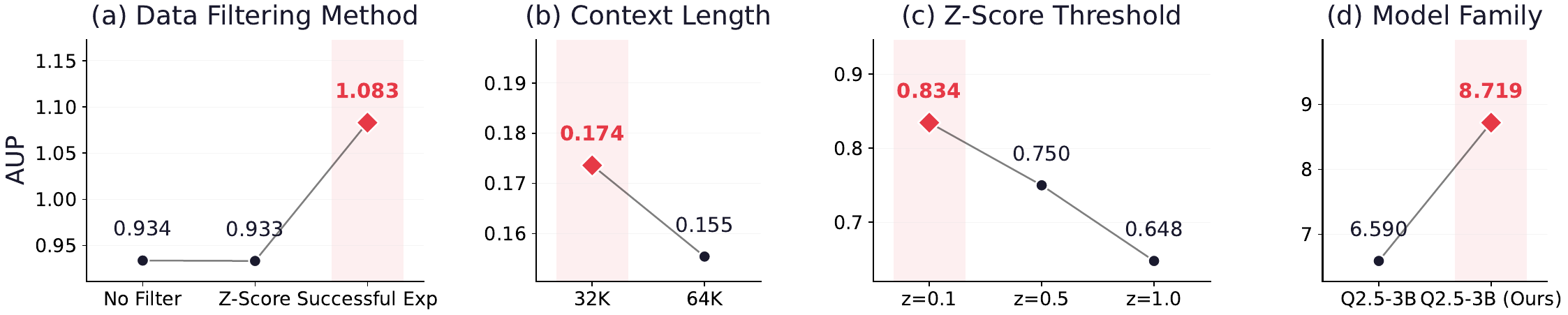}
    \caption{\textbf{Ablation Studies Comparison.} Performance comparison of ML-AutoResearch agents across different settings. Refer to \ref{appendix:ablation_details} for additional details.}
    \label{fig:abliation_res}
\end{figure}

% \begin{table}[t]
%     \centering
%     \begin{tabular}{lccc}
%         \toprule
%         Maximum cosine similarity & MLGym & MLAgentBench & Combined \\
%         \midrule
%         $>0.6$ & 6/555 (1.1\%) & 0/555 (0.0\%) & 6/555 (1.1\%) \\
%         $>0.7$ & 0/555 (0.0\%) & 0/555 (0.0\%) & 0/555 (0.0\%) \\
%         \bottomrule
%     \end{tabular}
%     \caption{\textbf{Semantic train--test similarity.} Number of synthetic tasks whose maximum cosine similarity to an evaluation task exceeds each threshold. Task descriptions are encoded with \texttt{nvidia/llama-embed-nemotron-8b}.}
%     \label{tab:contamination}
% \end{table}

\begin{table}[t]
    \centering
    \small
    \begin{NiceTabular}{lccc}
        \toprule
        Similarity threshold & MLGym & MLAgentBench & Combined \\
        \midrule
        $>0.6$ & 6/555 (1.1\%) & 0/555 (0.0\%) & 6/555 (1.1\%) \\
        $>0.7$ & 0/555 (0.0\%) & 0/555 (0.0\%) & 0/555 (0.0\%) \\
        \bottomrule
    \end{NiceTabular}
    \vspace{3mm}
    \caption{\textbf{Semantic train--test similarity.} Number of synthetic tasks whose maximum cosine similarity to an evaluation task exceeds each threshold. Task descriptions are encoded with \texttt{nvidia/llama-embed-nemotron-8b}.}
    \label{tab:contamination}
\end{table}

\subsection{Ablation Studies}
\label{sec:ablations}

% % We present ablation studies to quantify the impact of trajectory length constraints, data filtering strategies, and base model architectures on the efficacy of the ML-AutoResearch framework.

We present ablations on trajectory length, data filtering, and base model architectures to quantify their impact on the ML-AutoResearch framework.

\paragraph{Training Data Selection Mechanisms.}
To construct the final training dataset from candidate rollouts, we compared three selection mechanisms: (1) an unfiltered baseline, (2) Z-score sampling (discarding trajectories below a standardized performance threshold), and (3) successful experience selection (retaining only trajectories outperforming the baseline implementation). As Figure~\ref{fig:abliation_res}(a) shows, successful experience selection performs best by explicitly eliminating noisy, suboptimal executions. Z-score sampling improves upon the baseline but still admits some degenerate trajectories, highlighting the necessity of strict, performance-aligned curation.

\begin{wrapfigure}{r}{0.45\textwidth}
    \vspace{-8mm}
    \centering
    \includegraphics[width=0.4\textwidth]{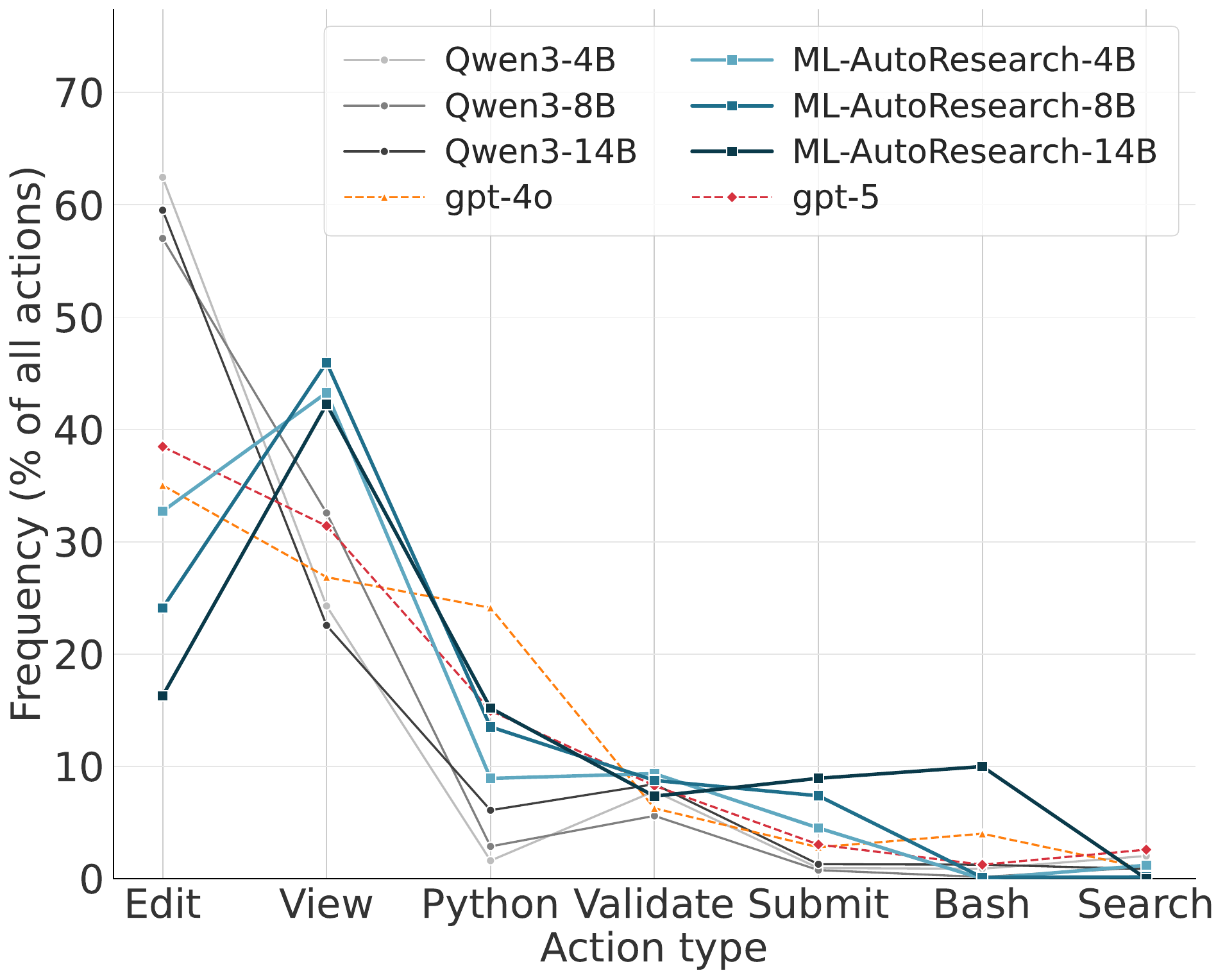}
    % \vspace{-7mm}
    \caption{\textbf{Distribution of Action Types.} Comparison of the distribution of different action types across agents.}
    % \vspace{-3mm}
    \label{fig:action_distribution}
    \vspace{-7mm}
\end{wrapfigure}

\paragraph{Trajectory Integrity vs. Dataset Size.}
Because ${\sim}35\%$ of raw trajectories exceed our 32,768-token limit, we compared training Qwen3-4B on strictly complete trajectories versus a dataset that truncates over-length rollouts. Despite providing fewer training samples, training exclusively on complete trajectories consistently outperformed the truncated variant (Figure~\ref{fig:abliation_res}(b)). Preserving full temporal and causal coherence is thus more critical for effective credit assignment than artificially inflating dataset size.

\paragraph{Sensitivity to Filtering Thresholds.}
Evaluating varying Z-score thresholds ($z \in \{0.1, 0.5, 1.0\}$) reveals that a stricter threshold of $z=0.1$ achieves the best empirical results (Figure~\ref{fig:abliation_res}(c)). Lowering the threshold increases dataset size but admits noisier trajectories, reinforcing that a smaller, high-signal dataset is vastly preferable for agentic fine-tuning.

\paragraph{Generalization Across Model Families.}
To confirm that ML-AutoResearch is model-agnostic, we applied our synthetic fine-tuning pipeline to an alternative architecture family, Qwen2.5-3B-Instruct. Evaluated on MLGym (Figure~\ref{fig:abliation_res} (d)), the ML-AR tuned variant demonstrated consistent and significant performance improvements over its base model. This establishes that our data generation and filtering approach effectively transfers across different base model families and parameter scales.

\subsection{Error Analysis}
\label{sec:error_analysis}
\begin{figure}[t]
    \centering
    \includegraphics[width=1\linewidth]{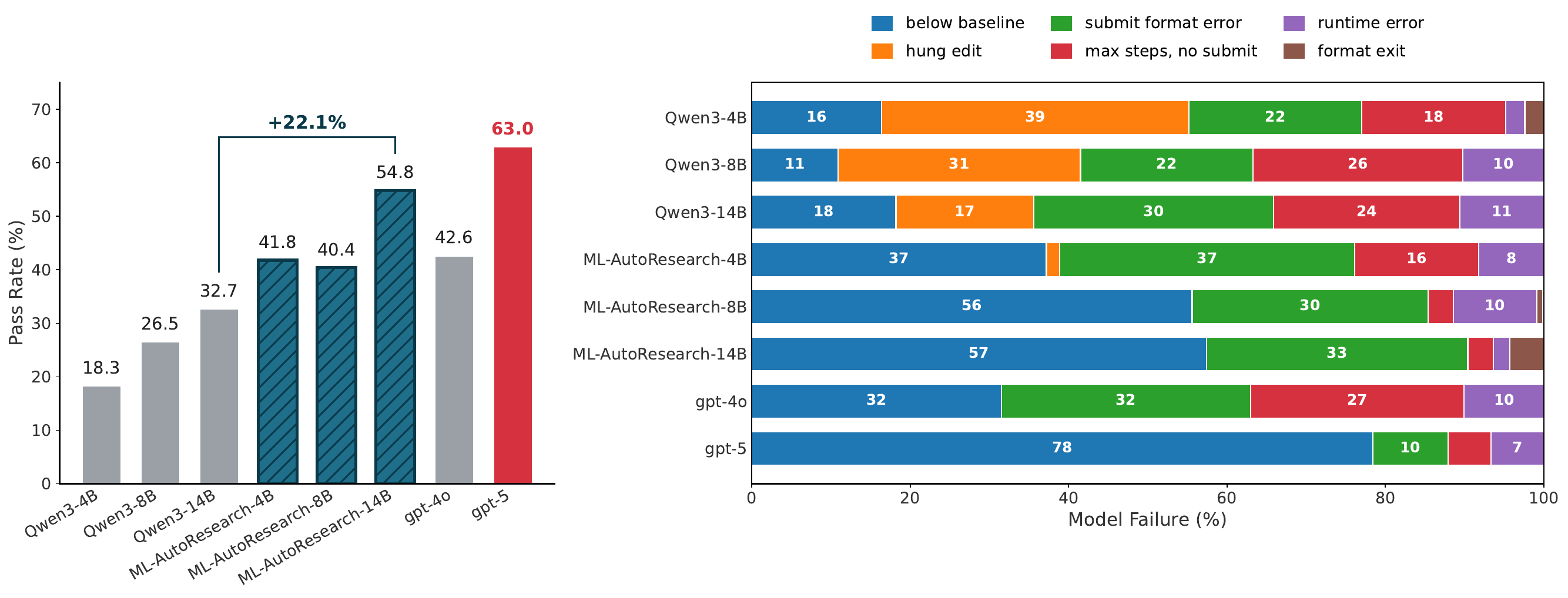}
    \vspace{-3mm}
    \caption{\textbf{Analysis Results.} Left: Comparison of ML-AutoResearch performance with other models based on pass rate, indicating how often agents outperform baseline results. Right: Overview of error categories across different agents.}
    \vspace{-3mm}
    \label{fig:error_categories}
\end{figure}

% To better understand the behavioral differences across models, we conducted a detailed failure analysis of the generated trajectories. As illustrated in Figure~\ref{fig:error_categories}, we categorize terminal failures into six distinct classes: \textit{Below Baseline}, \textit{Submit Format Error}, \textit{Runtime Error}, \textit{Hung Edit}, \textit{Max Steps Reached}, and \textit{Format Exit}. 
% Our analysis reveals several key insights regarding agent behavior:
To understand behavioral differences, we categorized terminal failures into six classes (Figure~\ref{fig:error_categories}). Baseline models predominantly fall into editing loops, repeatedly applying ineffective code changes until they exhaust their limits. In contrast, ML-AutoResearch tuned models successfully escape these degenerate loops to explore alternative strategies. Consequently, the primary failure mode for our agents shifts to submission format errors, reflecting occasional struggles with strict MLGym schemas.

While frontier models also avoid editing loops, they frequently fail by scoring below the baseline. This highlights a lack of exploratory creativity to design high-performing solutions, a bottleneck ML-AutoResearch-tuned models specifically overcome. Additionally, runtime errors from failing to debug written code constitute a small fraction of failures. The incidence of unrecovered runtime errors decreases further in our ML-AutoResearch-14B model, demonstrating that scaling the framework yields highly robust debugging capabilities.

\section{Conclusion}
We introduced ML-AutoResearch (ML-AR), a scalable framework for generating executable ML research tasks and process-level training trajectories. From 1,000 sampled topics, ML-AR produces 555 verified environments; 34K teacher trajectories are then filtered to 17K high-quality examples for supervised fine-tuning. Three-seed MLGym evaluation shows mean AUP gains of 5.7\%, 9.1\%, and 5.9\% at 4B, 8B, and 14B, respectively, while additional mixed GPT-5/GPT-5.4 data raises the 14B model's best-run AUP by 7.0\% over its instruction-tuned baseline. Improvements on MLAgentBench and AutoResearch further indicate that the learned process of experimentation, debugging, and feedback-driven refinement transfers beyond the primary evaluation environment.

\paragraph{Limitations}
Our training objective is supervised imitation, so ML-AR teaches feedback-driven execution and refinement rather than explicitly optimizing exploration or scientific novelty; reinforcement learning on the generated environments is a natural next step. The task-generation and verification stages remain mediated by GPT-5. Although an open-weight trajectory teacher also improves downstream performance (Appendix). Finally, generating and evaluating long-horizon trajectories requires substantial compute. Broader impacts are discussed in the appendix under the ``Broader impacts'' subsection.

\bibliographystyle{plainnat}
\newpage
\bibliography{references}

@article{zhang2025mlrc,
  title={MLRC-Bench: Can Language Agents Solve Machine Learning Research Challenges?},
  author={Zhang, Yunxiang and Khalifa, Muhammad and Bhushan, Shitanshu and Murphy, Lajanugen and Kim, Jaekyeom and Lee, Moontae and Lee, Honglak and Wang, Lu},
  journal={arXiv preprint arXiv:2504.09702},
  year={2025}
}

@article{lu2024ai,
  title={The AI Scientist: Towards Fully Automated Open-Ended Scientific Discovery},
  author={Lu, Chris and Lu, Cong and Lange, Robert Tjarko and Foerster, Jakob and Clune, Jeff and Ha, David},
  journal={arXiv preprint arXiv:2408.06292},
  year={2024}
}

@inproceedings{starace2025paperbench,
title={PaperBench: Evaluating {AI}{\textquoteright}s Ability to Replicate {{AI}} Research},
author={Giulio Starace and Oliver Jaffe and Dane Sherburn and James Aung and Jun Shern Chan and Leon Maksin and Rachel Dias and Evan Mays and Benjamin Kinsella and Wyatt Thompson and Johannes Heidecke and Amelia Glaese and Tejal Patwardhan},
booktitle={Forty-second International Conference on Machine Learning},
year={2025},
url={https://openreview.net/forum?id=xF5PuTLPbn}
}

@article{chen2025mlr,
  title={Mlr-bench: Evaluating ai agents on open-ended machine learning research},  author={Chen, Hui and Xiong, Miao and Lu, Yujie and Han, Wei and Deng, Ailin and He, Yufei and Wu, Jiaying and Li, Yibo and Liu, Yue and Hooi, Bryan},
  journal={arXiv preprint arXiv:2505.19955},
  year={2025}
}

@article{gottweis2025towards,
  title={Towards an AI co-scientist},
  author={Gottweis, Juraj and Weng, Wei-Hung and Daryin, Alexander and Tu, Tao and Palepu, Anil and Sirkovic, Petar and Myaskovsky, Artiom and Weissenberger, Felix and Rong, Keran and Tanno, Ryutaro and others},
  journal={arXiv preprint arXiv:2502.18864},
  year={2025}
}

@article{siegel2024core,
  title={Core-bench: Fostering the credibility of published research through a computational reproducibility agent benchmark},
  author={Siegel, Zachary S and Kapoor, Sayash and Nagdir, Nitya and Stroebl, Benedikt and Narayanan, Arvind},
  journal={arXiv preprint arXiv:2409.11363},
  year={2024}
}

@article{chan2024mle,
  title={Mle-bench: Evaluating machine learning agents on machine learning engineering},  author={Chan, Jun Shern and Chowdhury, Neil and Jaffe, Oliver and Aung, James and Sherburn, Dane and Mays, Evan and Starace, Giulio and Liu, Kevin and Maksin, Leon and Patwardhan, Tejal and others},
  journal={arXiv preprint arXiv:2410.07095},
  year={2024}
}

@article{si2025ideation,
  title={The ideation-execution gap: Execution outcomes of llm-generated versus human research ideas},
  author={Si, Chenglei and Hashimoto, Tatsunori and Yang, Diyi},
  journal={arXiv preprint arXiv:2506.20803},
  year={2025}
}

@article{zhao2025automated,
  title={The Automated LLM Speedrunning Benchmark: Reproducing NanoGPT Improvements},
  author={Zhao, Bingchen and Magka, Despoina and Jiang, Minqi and Li, Xian and Raileanu, Roberta and Shavrina, Tatiana and Gagnon-Audet, Jean-Christophe and Niu, Kelvin and Sodhani, Shagun and Shvartsman, Michael and others},
  journal={arXiv preprint arXiv:2506.22419},
  year={2025}
}

@article{nathani2025mlgym,
  title={Mlgym: A new framework and benchmark for advancing ai research agents},
  author={Nathani, Deepak and Madaan, Lovish and Roberts, Nicholas and Bashlykov, Nikolay and Menon, Ajay and Moens, Vincent and Budhiraja, Amar and Magka, Despoina and Vorotilov, Vladislav and Chaurasia, Gaurav and others},
  journal={arXiv preprint arXiv:2502.14499},
  year={2025}
}

@article{huang2023mlagentbench,
  title={Mlagentbench: Evaluating language agents on machine learning experimentation},
  author={Huang, Qian and Vora, Jian and Liang, Percy and Leskovec, Jure},
  journal={arXiv preprint arXiv:2310.03302},
  year={2023}
}

@article{o2025sparks,
  title={Sparks of science: Hypothesis generation using structured paper data},  author={O'Neill, Charles and Ghosal, Tirthankar and R{\u{a}}ileanu, Roberta and Walmsley, Mike and Bui, Thang and Schawinski, Kevin and Ciuc{\u{a}}, Ioana},
  journal={arXiv preprint arXiv:2504.12976},
  year={2025}
}

@article{yao2022react,
  title={React: Synergizing reasoning and acting in language models},
  author={Yao, Shunyu and Zhao, Jeffrey and Yu, Dian and Du, Nan and Shafran, Izhak and Narasimhan, Karthik and Cao, Yuan},
  journal={arXiv preprint arXiv:2210.03629},
  year={2022}
}

@article{schick2023toolformer,
  title={Toolformer: Language models can teach themselves to use tools},
  author={Schick, Timo and Dwivedi-Yu, Jane and Dess{\`\i}, Roberto and Raileanu, Roberta and Lomeli, Maria and Hambro, Eric and Zettlemoyer, Luke and Cancedda, Nicola and Scialom, Thomas},
  journal={Advances in neural information processing systems},
  volume={36},
  pages={68539--68551},
  year={2023}
}

@article{yang2025swe,
  title={Swe-smith: Scaling data for software engineering agents},
  author={Yang, John and Lieret, Kilian and Jimenez, Carlos E and Wettig, Alexander and Khandpur, Kabir and Zhang, Yanzhe and Hui, Binyuan and Press, Ofir and Schmidt, Ludwig and Yang, Diyi},
  journal={arXiv preprint arXiv:2504.21798},
  year={2025}
}

@article{novikov2025alphaevolve,
  title={Alphaevolve: A coding agent for scientific and algorithmic discovery},
  author={Novikov, Alexander and V{\~u}, Ng{\^a}n and Eisenberger, Marvin and Dupont, Emilien and Huang, Po-Sen and Wagner, Adam Zsolt and Shirobokov, Sergey and Kozlovskii, Borislav and Ruiz, Francisco JR and Mehrabian, Abbas and others},
  journal={arXiv preprint arXiv:2506.13131},
  year={2025}
}

@article{yang2025qwen3,
  title={Qwen3 technical report},
  author={Yang, An and Li, Anfeng and Yang, Baosong and Zhang, Beichen and Hui, Binyuan and Zheng, Bo and Yu, Bowen and Gao, Chang and Huang, Chengen and Lv, Chenxu and others},
  journal={arXiv preprint arXiv:2505.09388},
  year={2025}
}

@article{yang2024swe,
  title={Swe-agent: Agent-computer interfaces enable automated software engineering},
  author={Yang, John and Jimenez, Carlos E and Wettig, Alexander and Lieret, Kilian and Yao, Shunyu and Narasimhan, Karthik and Press, Ofir},
  journal={Advances in Neural Information Processing Systems},
  volume={37},
  pages={50528--50652},
  year={2024}
}

@article{brown2020language,
  title={Language models are few-shot learners},
  author={Brown, Tom and Mann, Benjamin and Ryder, Nick and Subbiah, Melanie and Kaplan, Jared D and Dhariwal, Prafulla and Neelakantan, Arvind and Shyam, Pranav and Sastry, Girish and Askell, Amanda and others},
  journal={Advances in neural information processing systems},
  volume={33},
  pages={1877--1901},
  year={2020}
}

@article{touvron2023llama,
  title={Llama: Open and efficient foundation language models},
  author={Touvron, Hugo and Lavril, Thibaut and Izacard, Gautier and Martinet, Xavier and Lachaux, Marie-Anne and Lacroix, Timoth{\'e}e and Rozi{\`e}re, Baptiste and Goyal, Naman and Hambro, Eric and Azhar, Faisal and others},
  journal={arXiv preprint arXiv:2302.13971},
  year={2023}
}

\section*{NeurIPS Paper Checklist}

\begin{enumerate}

\item {\bf Claims}
    \item[] Question: Do the main claims made in the abstract and introduction accurately reflect the paper's contributions and scope?
    \item[] Answer: \answerYes{}
    \item[] Justification: The abstract and introduction state the core contribution (ML-AutoResearch), its setting, and the reported evaluation gains, and these are aligned with the method and experiments sections. See the abstract, Section~1, Section~\ref{sec:method}, and Section~4.

\item {\bf Limitations}
    \item[] Question: Does the paper discuss the limitations of the work performed by the authors?
    \item[] Answer: \answerYes{}
    \item[] Justification: A dedicated \textit{Limitations} paragraph is included in the conclusion, discussing supervision method constraints, dependence on a single teacher model, and open scaling questions. See Section~5.

\item {\bf Theory assumptions and proofs}
    \item[] Question: For each theoretical result, does the paper provide the full set of assumptions and a complete (and correct) proof?
    \item[] Answer: \answerNA{}
    \item[] Justification: The paper is primarily an empirical systems and benchmark study and does not present theorem-proof style theoretical claims requiring formal proofs.

    \item {\bf Experimental result reproducibility}
    \item[] Question: Does the paper fully disclose all the information needed to reproduce the main experimental results of the paper to the extent that it affects the main claims and/or conclusions of the paper (regardless of whether the code and data are provided or not)?
    \item[] Answer: \answerYes{}
    \item[] Justification: The appendix now specifies the training pipeline settings, model/training hyperparameters, sequence-length filtering, and trajectory selection knobs used in SFT. See Section~\ref{sec:appendix-repro}.

\item {\bf Open access to data and code}
    \item[] Question: Does the paper provide open access to the data and code, with sufficient instructions to faithfully reproduce the main experimental results, as described in supplemental material?
    \item[] Answer: \answerNo{}
    \item[] Justification: The current manuscript does not yet provide anonymized public repository links or release artifacts with full reproduction instructions in the supplement.

\item {\bf Experimental setting/details}
    \item[] Question: Does the paper specify all the training and test details (e.g., data splits, hyperparameters, how they were chosen, type of optimizer) necessary to understand the results?
    \item[] Answer: \answerYes{}
    \item[] Justification: The appendix enumerates optimizer, LR schedule, warmup, weight decay, batch/micro-batch, max steps, seed, sequence constraints, and filtering settings used in training. See Section~\ref{sec:appendix-repro}.

\item {\bf Experiment statistical significance}
    \item[] Question: Does the paper report error bars suitably and correctly defined or other appropriate information about the statistical significance of the experiments?
    \item[] Answer: \answerNo{}
    \item[] Justification: Repeating full training runs multiple times is not feasible under our compute budget, which is common for large-scale agent training setups. We therefore report full task-level outcomes and aggregate benchmark metrics, and we observe that evaluation is consistent across reruns; however, we do not provide hypothesis tests, so we conservatively mark this item as \answerNo{}.

\item {\bf Experiments compute resources}
    \item[] Question: For each experiment, does the paper provide sufficient information on the computer resources (type of compute workers, memory, time of execution) needed to reproduce the experiments?
    \item[] Answer: \answerYes{}
    \item[] Justification: The appendix lists the H100 cluster SKU, GPU counts, parallelism settings, and run-step/time-budget configuration used by training jobs. See Section~\ref{sec:appendix-repro}.
    
\item {\bf Code of ethics}
    \item[] Question: Does the research conducted in the paper conform, in every respect, with the NeurIPS Code of Ethics \url{https://neurips.cc/public/EthicsGuidelines}?
    \item[] Answer: \answerYes{}
    \item[] Justification: The work is conducted as benchmarked ML systems research using established public resources, and no deviations from the NeurIPS Code of Ethics are identified.

\item {\bf Broader impacts}
    \item[] Question: Does the paper discuss both potential positive societal impacts and negative societal impacts of the work performed?
    \item[] Answer: \answerYes{}
    \item[] Justification: The appendix includes a dedicated broader-impact section that covers positive impacts, negative misuse risks, and mitigation context for controlled evaluation settings (Section~\ref{sec:appendix-broader-impacts}).
    
\item {\bf Safeguards}
    \item[] Question: Does the paper describe safeguards that have been put in place for responsible release of data or models that have a high risk for misuse (e.g., pre-trained language models, image generators, or scraped datasets)?
    \item[] Answer: \answerNA{}
    \item[] Justification: The manuscript does not currently release high-risk generative model artifacts or scraped datasets that require special controlled-release safeguards.

\item {\bf Licenses for existing assets}
    \item[] Question: Are the creators or original owners of assets (e.g., code, data, models), used in the paper, properly credited and are the license and terms of use explicitly mentioned and properly respected?
    \item[] Answer: \answerYes{}
    \item[] Justification: We include an explicit appendix inventory of external assets and their license/terms categories (datasets, open models, API models, benchmarks/codebases, and tools), and state that usage follows those terms. See Section~\ref{sec:appendix-license-audit}.

\item {\bf New assets}
    \item[] Question: Are new assets introduced in the paper well documented and is the documentation provided alongside the assets?
    \item[] Answer: \answerNo{}
    \item[] Justification: The paper introduces new synthetic tasks and trajectories, but release-time documentation package details (e.g., datasheet-style metadata and access instructions) are not yet finalized in this draft.

\item {\bf Crowdsourcing and research with human subjects}
    \item[] Question: For crowdsourcing experiments and research with human subjects, does the paper include the full text of instructions given to participants and screenshots, if applicable, as well as details about compensation (if any)? 
    \item[] Answer: \answerNA{}
    \item[] Justification: This work does not involve crowdsourcing studies or research with human participants.

\item {\bf Institutional review board (IRB) approvals or equivalent for research with human subjects}
    \item[] Question: Does the paper describe potential risks incurred by study participants, whether such risks were disclosed to the subjects, and whether Institutional Review Board (IRB) approvals (or an equivalent approval/review based on the requirements of your country or institution) were obtained?
    \item[] Answer: \answerNA{}
    \item[] Justification: The study does not include human-subject data collection or interventions requiring IRB or equivalent review.

\item {\bf Declaration of LLM usage}
    \item[] Question: Does the paper describe the usage of LLMs if it is an important, original, or non-standard component of the core methods in this research? Note that if the LLM is used only for writing, editing, or formatting purposes and does \emph{not} impact the core methodology, scientific rigor, or originality of the research, declaration is not required.
    %this research? 
    \item[] Answer: \answerYes{}
    \item[] Justification: LLM usage is central to the method: a teacher model is used for task synthesis, verification, and trajectory generation, and this is explicitly described throughout Section~\ref{sec:method} and the experiments.

\end{enumerate}
\appendix
\section{Appendix}
\subsection{Benchmarks Details}

\label{sec:appendix_benchmarks_details}

\paragraph{MLGym benchmark (13 subtasks).}
We evaluate on the 13-task MLGym suite spanning supervised learning, self-supervised
learning, reinforcement learning, algorithmic reasoning, and game theory. The subtasks are:
\textit{House Price Prediction}, \textit{Image Classification (CIFAR-10)},
\textit{Image Classification (Fashion-MNIST)}, \textit{Image Captioning (MS-COCO)},
\textit{Natural Language Inference (MNLI)}, \textit{Language Modeling (FineWeb)},
\textit{MetaMaze Navigation}, \textit{MountainCar Continuous},
\textit{Breakout MinAtar}, \textit{3-SAT}, \textit{Prisoner's Dilemma},
\textit{Battle of Sexes}, and \textit{Colonel Blotto}.

Each MLGym subtask uses its task-native evaluator to produce a scalar score, with
directionality (maximize/minimize) defined per task by the benchmark configuration.
Following benchmark convention, we report both (i) whether an agent beats the provided
baseline (pass-style success) and (ii) aggregate trajectory quality via AUP over
intermediate submissions.

\paragraph{MLAgentBench benchmark (reported subtasks and metrics).}
For MLAgentBench, we report the following subtasks and their native evaluation metrics:
\textit{CIFAR-10} (Accuracy, $\uparrow$), \textit{IMDb} (Accuracy, $\uparrow$),
\textit{House Price Prediction} (MAE, $\downarrow$), \textit{ogbn-arxiv}
(Accuracy, $\uparrow$), \textit{Vectorization} (runtime sec, $\downarrow$),
\textit{Spaceship Titanic} (Accuracy, $\uparrow$), \textit{AMP Parkinsons}
(SMAPE, $\downarrow$), \textit{LLaMA Inference} (sec/token, $\downarrow$), and
\textit{Feedback Prize} (MCRMSE, $\downarrow$). Metrics are computed by the task
evaluator and compared by direction-aware best-attempt performance.

\subsection{Multi-Seed Evaluation Details}
\label{sec:appendix-multiseed-details}
Table~\ref{tab:mlgym_per_task_multiseed} reports the complete per-task MLGym results for the primary three-seed reevaluation. ML-AR generally improves execution reliability, converting many invalid base-model runs into valid solutions, while outcomes remain heterogeneous: for example, the 4B MNLI and 14B Breakout results regress. Table~\ref{tab:mlagentbench_per_task_multiseed} reports task-level three-seed MLAgentBench results, and Table~\ref{tab:mlagentbench_multiseed_seeds} provides the seed-level normalized scores underlying the aggregate results in Table~\ref{tab:mlagentbench_multiseed}.

\begin{table*}[t]
    \centering
    \scriptsize
    \setlength{\tabcolsep}{2.5pt}
    \begin{adjustbox}{width=\textwidth}
    \begin{NiceTabular}{llcccccc}
        \toprule
        Task & Metric & Qwen3-4B & ML-AR-4B & Qwen3-8B & ML-AR-8B & Qwen3-14B & ML-AR-14B \\
        \midrule

        CIFAR-10 & Accuracy ($\uparrow$)
        & $0.000_{\scriptscriptstyle \pm 0.000}$ [0/3]
        & $0.000_{\scriptscriptstyle \pm 0.000}$ [0/3]
        & $0.000_{\scriptscriptstyle \pm 0.000}$ [0/3]
        & $0.667_{\scriptscriptstyle \pm 0.577}$ [2/3]
        & $0.243_{\scriptscriptstyle \pm 0.421}$ [1/3]
        & $0.667_{\scriptscriptstyle \pm 0.577}$ [2/3] \\

        Battle of Sexes & Avg. reward ($\uparrow$)
        & $1.445_{\scriptscriptstyle \pm 0.002}$ [3/3]
        & $1.804_{\scriptscriptstyle \pm 0.002}$ [3/3]
        & $1.445_{\scriptscriptstyle \pm 0.002}$ [3/3]
        & $1.759_{\scriptscriptstyle \pm 0.265}$ [3/3]
        & $1.446_{\scriptscriptstyle \pm 0.001}$ [3/3]
        & $1.870_{\scriptscriptstyle \pm 0.113}$ [3/3] \\

        Prisoners Dilemma & Avg. reward ($\uparrow$)
        & $2.629_{\scriptscriptstyle \pm 0.005}$ [3/3]
        & $3.388_{\scriptscriptstyle \pm 0.002}$ [3/3]
        & $2.624_{\scriptscriptstyle \pm 0.010}$ [3/3]
        & $2.635_{\scriptscriptstyle \pm 0.002}$ [3/3]
        & $2.633_{\scriptscriptstyle \pm 0.003}$ [3/3]
        & $2.638_{\scriptscriptstyle \pm 0.003}$ [3/3] \\

        Blotto & Avg. reward ($\uparrow$)
        & $0.208_{\scriptscriptstyle \pm 0.074}$ [3/3]
        & $0.662_{\scriptscriptstyle \pm 0.473}$ [3/3]
        & $0.096_{\scriptscriptstyle \pm 0.129}$ [3/3]
        & $0.334_{\scriptscriptstyle \pm 0.577}$ [3/3]
        & $0.340_{\scriptscriptstyle \pm 0.443}$ [3/3]
        & $0.111_{\scriptscriptstyle \pm 0.126}$ [3/3] \\

        House Price & $R^2$ ($\uparrow$)
        & $0.596_{\scriptscriptstyle \pm 0.517}$ [2/3]
        & $0.880_{\scriptscriptstyle \pm 0.000}$ [3/3]
        & $0.899_{\scriptscriptstyle \pm 0.010}$ [3/3]
        & $0.914_{\scriptscriptstyle \pm 0.002}$ [3/3]
        & $0.908_{\scriptscriptstyle \pm 0.011}$ [3/3]
        & $0.918_{\scriptscriptstyle \pm 0.002}$ [3/3] \\

        Fashion MNIST & Accuracy ($\uparrow$)
        & $0.000_{\scriptscriptstyle \pm 0.000}$ [0/3]
        & $0.000_{\scriptscriptstyle \pm 0.000}$ [0/3]
        & $0.000_{\scriptscriptstyle \pm 0.000}$ [0/3]
        & $0.672_{\scriptscriptstyle \pm 0.497}$ [3/3]
        & $0.000_{\scriptscriptstyle \pm 0.000}$ [0/3]
        & $0.977_{\scriptscriptstyle \pm 0.040}$ [3/3] \\

        MS-COCO & BLEU ($\uparrow$)
        & $0.008_{\scriptscriptstyle \pm 0.014}$ [1/3]
        & $0.280_{\scriptscriptstyle \pm 0.000}$ [3/3]
        & $0.323_{\scriptscriptstyle \pm 0.006}$ [3/3]
        & $0.535_{\scriptscriptstyle \pm 0.404}$ [3/3]
        & $0.297_{\scriptscriptstyle \pm 0.018}$ [3/3]
        & $0.760_{\scriptscriptstyle \pm 0.416}$ [3/3] \\

        MNLI & Accuracy ($\uparrow$)
        & $0.834_{\scriptscriptstyle \pm 0.004}$ [3/3]
        & $0.537_{\scriptscriptstyle \pm 0.008}$ [3/3]
        & $0.000_{\scriptscriptstyle \pm 0.000}$ [0/3]
        & $0.766_{\scriptscriptstyle \pm 0.008}$ [3/3]
        & $0.557_{\scriptscriptstyle \pm 0.482}$ [2/3]
        & $0.735_{\scriptscriptstyle \pm 0.178}$ [3/3] \\

        Language Modeling & Val. loss ($\downarrow$)
        & $4.924_{\scriptscriptstyle \pm 0.006}$ [2/3]
        & $4.922_{\scriptscriptstyle \pm 0.001}$ [3/3]
        & $4.931_{\scriptscriptstyle \pm 0.000}$ [1/3]
        & $4.919_{\scriptscriptstyle \pm 0.004}$ [3/3]
        & $4.931_{\scriptscriptstyle \pm 0.000}$ [0/3]
        & $4.917_{\scriptscriptstyle \pm 0.002}$ [3/3] \\

        Breakout & Avg. score ($\uparrow$)
        & $0.000_{\scriptscriptstyle \pm 0.000}$ [0/3]
        & $41.233_{\scriptscriptstyle \pm 7.447}$ [3/3]
        & $0.000_{\scriptscriptstyle \pm 0.000}$ [0/3]
        & $29.970_{\scriptscriptstyle \pm 18.123}$ [3/3]
        & $48.137_{\scriptscriptstyle \pm 5.383}$ [3/3]
        & $39.793_{\scriptscriptstyle \pm 1.396}$ [3/3] \\

        Mountain Car & Avg. reward ($\uparrow$)
        & $0.000_{\scriptscriptstyle \pm 0.000}$ [0/3]
        & $49.830_{\scriptscriptstyle \pm 1.923}$ [3/3]
        & $16.240_{\scriptscriptstyle \pm 28.129}$ [1/3]
        & $52.050_{\scriptscriptstyle \pm 0.000}$ [3/3]
        & $29.321_{\scriptscriptstyle \pm 23.829}$ [3/3]
        & $50.940_{\scriptscriptstyle \pm 1.923}$ [3/3] \\

        Meta Maze & Avg. return ($\uparrow$)
        & $0.000_{\scriptscriptstyle \pm 0.000}$ [0/3]
        & $21.203_{\scriptscriptstyle \pm 3.837}$ [3/3]
        & $8.327_{\scriptscriptstyle \pm 14.422}$ [1/3]
        & $21.800_{\scriptscriptstyle \pm 2.803}$ [3/3]
        & $5.394_{\scriptscriptstyle \pm 4.680}$ [2/3]
        & $21.433_{\scriptscriptstyle \pm 3.223}$ [3/3] \\

        3-SAT & Time (s) ($\downarrow$)
        & $13.397_{\scriptscriptstyle \pm 0.578}$ [3/3]
        & $13.740_{\scriptscriptstyle \pm 0.667}$ [3/3]
        & $13.663_{\scriptscriptstyle \pm 0.356}$ [3/3]
        & $13.930_{\scriptscriptstyle \pm 0.451}$ [3/3]
        & $11.620_{\scriptscriptstyle \pm 0.061}$ [3/3]
        & $12.920_{\scriptscriptstyle \pm 0.713}$ [3/3] \\

        \bottomrule
    \end{NiceTabular}
    \end{adjustbox}
    \caption{\textbf{Per-task three-seed MLGym results.} Values are mean and standard deviation across evaluation seeds 42, 123, and 456; brackets show the number of valid runs. Invalid runs follow the benchmark aggregation convention used in the rebuttal results. The 14B column evaluates the original ML-AR-14B checkpoint, not ML-AR-14B-Mixed.}
    \label{tab:mlgym_per_task_multiseed}
\end{table*}
% \begin{table*}[t]
%     \centering
%     \scriptsize
%     \setlength{\tabcolsep}{3pt}
%     \begin{adjustbox}{width=\textwidth}
%     \begin{tabular}{lcccccc}
%         \toprule
%         Task & Qwen3-4B & ML-AR-4B & Qwen3-8B & ML-AR-8B & Qwen3-14B & ML-AR-14B \\
%         \midrule
%         CIFAR-10      & $0.000 \pm 0.000$ & $0.171 \pm 0.296$ & $0.221 \pm 0.383$ & $0.418 \pm 0.363$ & $0.274 \pm 0.392$ & $0.867 \pm 0.010$ \\
%         IMDb          & $0.000 \pm 0.000$ & $0.000 \pm 0.000$ & $0.000 \pm 0.000$ & $0.304 \pm 0.527$ & $0.000 \pm 0.000$ & $0.611 \pm 0.529$ \\
%         ogbn-arxiv    & $0.000 \pm 0.000$ & $0.000 \pm 0.000$ & $0.000 \pm 0.000$ & $0.350 \pm 0.304$ & $0.357 \pm 0.310$ & $0.431 \pm 0.221$ \\
%         Vectorization & $0.000 \pm 0.000$ & $0.334 \pm 0.577$ & $0.001 \pm 0.001$ & $0.001 \pm 0.002$ & $0.000 \pm 0.000$ & $0.192 \pm 0.332$ \\
%         BabyLM-fixed  & -- & -- & -- & -- & -- & $6.03 \pm 0.98$ [2/3]$^{\dagger}$ \\
%         \bottomrule
%     \end{tabular}
%     \end{adjustbox}
%     \caption{\textbf{Task-level three-seed MLAgentBench results.} Values are mean $\pm$ standard deviation across evaluation seeds 42, 123, and 456. A dash indicates that the model did not solve BabyLM-fixed. $^{\dagger}$BabyLM-fixed reports evaluation loss (lower is better) and was solved in two of three runs; the other task columns report normalized performance (higher is better).}
%     \label{tab:mlagentbench_per_task_multiseed}
% \end{table*}

\begin{table*}[t]
    \centering
    \scriptsize
    \setlength{\tabcolsep}{3pt}
    \begin{adjustbox}{width=\textwidth}
    \begin{NiceTabular}{lcccccc}
        \toprule
        Task & Qwen3-4B & ML-AR-4B & Qwen3-8B & ML-AR-8B & Qwen3-14B & ML-AR-14B \\
        \midrule

        CIFAR-10
        & $0.000_{\scriptscriptstyle \pm 0.000}$
        & $0.171_{\scriptscriptstyle \pm 0.296}$
        & $0.221_{\scriptscriptstyle \pm 0.383}$
        & $0.418_{\scriptscriptstyle \pm 0.363}$
        & $0.274_{\scriptscriptstyle \pm 0.392}$
        & $0.867_{\scriptscriptstyle \pm 0.010}$ \\

        IMDb
        & $0.000_{\scriptscriptstyle \pm 0.000}$
        & $0.000_{\scriptscriptstyle \pm 0.000}$
        & $0.000_{\scriptscriptstyle \pm 0.000}$
        & $0.304_{\scriptscriptstyle \pm 0.527}$
        & $0.000_{\scriptscriptstyle \pm 0.000}$
        & $0.611_{\scriptscriptstyle \pm 0.529}$ \\

        ogbn-arxiv
        & $0.000_{\scriptscriptstyle \pm 0.000}$
        & $0.000_{\scriptscriptstyle \pm 0.000}$
        & $0.000_{\scriptscriptstyle \pm 0.000}$
        & $0.350_{\scriptscriptstyle \pm 0.304}$
        & $0.357_{\scriptscriptstyle \pm 0.310}$
        & $0.431_{\scriptscriptstyle \pm 0.221}$ \\

        Vectorization
        & $0.000_{\scriptscriptstyle \pm 0.000}$
        & $0.334_{\scriptscriptstyle \pm 0.577}$
        & $0.001_{\scriptscriptstyle \pm 0.001}$
        & $0.001_{\scriptscriptstyle \pm 0.002}$
        & $0.000_{\scriptscriptstyle \pm 0.000}$
        & $0.192_{\scriptscriptstyle \pm 0.332}$ \\

        BabyLM-fixed
        & $\infty$
        & $\infty$
        & $\infty$
        & $\infty$
        & $\infty$
        & $6.03_{\scriptscriptstyle \pm 0.98}$ [2/3]$^{\dagger}$ \\

        \bottomrule
    \end{NiceTabular}
    \end{adjustbox}
    \caption{\textbf{Task-level three-seed MLAgentBench results.} Values are mean and standard deviation across evaluation seeds 42, 123, and 456. A dash indicates that the model did not solve BabyLM-fixed. $^{\dagger}$BabyLM-fixed reports evaluation loss (lower is better) and was solved in two of three runs; the other task columns report normalized performance (higher is better).}
    \label{tab:mlagentbench_per_task_multiseed}
\end{table*}
% \begin{table}[t]
%     \centering
%     \begin{tabular}{lcccc}
%         \toprule
%         Model & Mean normalized score & Seed 42 & Seed 123 & Seed 456 \\
%         \midrule
%         Qwen3-4B  & $0.000 \pm 0.000$ & 0.000 & 0.000 & 0.000 \\
%         ML-AR-4B  & $0.101 \pm 0.100$ & 0.103 & 0.000 & 0.200 \\
%         Qwen3-8B  & $0.044 \pm 0.077$ & 0.000 & 0.133 & 0.000 \\
%         ML-AR-8B  & $0.215 \pm 0.212$ & 0.425 & 0.219 & 0.000 \\
%         Qwen3-14B & $0.126 \pm 0.127$ & 0.124 & 0.000 & 0.254 \\
%         ML-AR-14B & $0.420 \pm 0.193$ & 0.207 & 0.469 & 0.584 \\
%         \bottomrule
%     \end{tabular}
%     \caption{\textbf{MLAgentBench results by evaluation seed.} Mean normalized score and standard deviation across seeds 42, 123, and 456, together with each seed-level score.}
%     \label{tab:mlagentbench_multiseed_seeds}
% \end{table}

\begin{table}[t]
    \centering
    \small
    \setlength{\tabcolsep}{4pt}
    \begin{NiceTabular}{lcccc}
        \toprule
        Model & Mean score & \multicolumn{3}{c}{Per-seed score} \\
        \cmidrule(lr){3-5}
              &            & 42 & 123 & 456 \\
        \midrule

        Qwen3-4B
        & $0.000_{\scriptscriptstyle \pm 0.000}$
        & 0.000 & 0.000 & 0.000 \\

        ML-AR-4B
        & $0.101_{\scriptscriptstyle \pm 0.100}$
        & 0.103 & 0.000 & 0.200 \\

        Qwen3-8B
        & $0.044_{\scriptscriptstyle \pm 0.077}$
        & 0.000 & 0.133 & 0.000 \\

        ML-AR-8B
        & $0.215_{\scriptscriptstyle \pm 0.212}$
        & 0.425 & 0.219 & 0.000 \\

        Qwen3-14B
        & $0.126_{\scriptscriptstyle \pm 0.127}$
        & 0.124 & 0.000 & 0.254 \\

        ML-AR-14B
        & $0.420_{\scriptscriptstyle \pm 0.193}$
        & 0.207 & 0.469 & 0.584 \\

        \bottomrule
    \end{NiceTabular}
    \vspace{3mm}
    \caption{\textbf{MLAgentBench results by evaluation seed.} Mean normalized score and standard deviation across evaluation seeds 42, 123, and 456, together with each seed-level score.}
    \label{tab:mlagentbench_multiseed_seeds}
\end{table}

\subsection{Robustness Across Training Seeds}
\label{sec:appendix-training-seeds}
We independently train the 8B and 14B models using seeds 42, 1234, and 5678 and evaluate each resulting checkpoint with three independent evaluation seeds. Table~\ref{tab:training_seed_robustness} reports mean AUP gains of 4.4\% at 8B and 4.0\% at 14B. This protocol measures sensitivity to training randomness and is distinct from the primary-checkpoint evaluation in Table~\ref{tab:mlgym_multiseed}. The 14B checkpoints use the original ML-AR dataset rather than the mixed GPT-5/GPT-5.4 dataset.

% \begin{table}[t]
%     \centering
%     \begin{tabular}{lccc}
%         \toprule
%         Model size & Qwen3-Instruct AUP & ML-AR AUP & Relative gain \\
%         \midrule
%         8B  & $0.871 \pm 0.012$ & $0.910 \pm 0.040$ & $+4.4\%$ \\
%         14B & $0.904 \pm 0.025$ & $0.940 \pm 0.035$ & $+4.0\%$ \\
%         \bottomrule
%     \end{tabular}
%     \caption{\textbf{Robustness across training seeds.} Mean AUP across independently trained checkpoints using seeds 42, 1234, and 5678, with each checkpoint evaluated using three independent evaluation seeds. The 14B result uses the original ML-AR dataset rather than the mixed-data model.}
%     \label{tab:training_seed_robustness}
% \end{table}

\begin{table}[t]
    \centering
    \small
    \begin{NiceTabular}{lccc}
        \toprule
        Model & Qwen3 & ML-AR & Rel. gain \\
        \midrule
        8B
        & $0.871_{\scriptscriptstyle \pm 0.012}$
        & $0.910_{\scriptscriptstyle \pm 0.040}$
        & \textbf{$+4.4\%$} \\
        
        14B
        & $0.904_{\scriptscriptstyle \pm 0.025}$
        & $0.940_{\scriptscriptstyle \pm 0.035}$
        & \textbf{$+4.0\%$} \\
        \bottomrule
    \end{NiceTabular}
    \vspace{3mm}
    \caption{\textbf{Robustness across training seeds.} Mean AUP across independently trained checkpoints using seeds 42, 1234, and 5678, with each checkpoint evaluated using three independent evaluation seeds. The 14B result uses the original ML-AR dataset rather than the mixed-data model.}
    \label{tab:training_seed_robustness}
\end{table}

\subsection{Open-Weight Trajectory Teacher}
\label{sec:appendix-open-weight-teacher}
To test whether the trajectory-generation stage depends specifically on GPT-5 behavior, we replace GPT-5 with the open-weight Qwen3-32B model while holding the 555 verified task environments fixed. This experiment evaluates trajectory-teacher transfer only; it does not regenerate topics, datasets, or task environments with Qwen3-32B. The model generated 6K trajectories, of which approximately 3.9K were removed by the filtering pipeline, leaving 2.1K supervised fine-tuning examples. The aggregate result is reported in Table~\ref{tab:open_weight_teacher} in the main text. Table~\ref{tab:open_weight_teacher_per_task} gives the corresponding task-level results and valid-run counts, showing heterogeneous transfer despite positive aggregate gains. This provides initial evidence that downstream gains are not specific to GPT-5-generated trajectories, while leaving full open-weight environment synthesis for future work.

\begin{table*}[t]
    \centering
    \scriptsize
    \setlength{\tabcolsep}{3pt}
    \begin{adjustbox}{width=\textwidth}
    \begin{NiceTabular}{llcccc}
        \toprule
        Task & Metric & Qwen3-8B & ML-AR-8B & Qwen3-14B & ML-AR-14B \\
        \midrule

        CIFAR-10 & Accuracy ($\uparrow$)
        & $\infty$ [0/3]
        & $\infty$ [0/3]
        & $0.730_{\scriptscriptstyle \pm 0.000}$ [1/3]
        & $0.733_{\scriptscriptstyle \pm 0.000}$ [1/3] \\

        Battle of Sexes & Avg. reward ($\uparrow$)
        & $1.445_{\scriptscriptstyle \pm 0.001}$ [3/3]
        & $1.424_{\scriptscriptstyle \pm 0.315}$ [3/3]
        & $1.445_{\scriptscriptstyle \pm 0.001}$ [3/3]
        & $1.804_{\scriptscriptstyle \pm 0.000}$ [3/3] \\

        Prisoners Dilemma & Avg. reward ($\uparrow$)
        & $2.624_{\scriptscriptstyle \pm 0.008}$ [3/3]
        & $2.802_{\scriptscriptstyle \pm 0.424}$ [3/3]
        & $2.633_{\scriptscriptstyle \pm 0.003}$ [3/3]
        & $2.971_{\scriptscriptstyle \pm 0.422}$ [3/3] \\

        Blotto & Avg. reward ($\uparrow$)
        & $0.096_{\scriptscriptstyle \pm 0.105}$ [3/3]
        & $0.227_{\scriptscriptstyle \pm 0.506}$ [3/3]
        & $0.340_{\scriptscriptstyle \pm 0.362}$ [3/3]
        & $0.199_{\scriptscriptstyle \pm 0.049}$ [3/3] \\

        House Price & $R^2$ ($\uparrow$)
        & $0.899_{\scriptscriptstyle \pm 0.007}$ [3/3]
        & $0.880_{\scriptscriptstyle \pm 0.000}$ [3/3]
        & $0.908_{\scriptscriptstyle \pm 0.009}$ [3/3]
        & $0.900_{\scriptscriptstyle \pm 0.001}$ [3/3] \\

        Fashion MNIST & Accuracy ($\uparrow$)
        & $\infty$ [0/3]
        & $\infty$ [0/3]
        & $\infty$ [0/3]
        & $0.949_{\scriptscriptstyle \pm 0.000}$ [1/3] \\

        MS-COCO & BLEU ($\uparrow$)
        & $0.323_{\scriptscriptstyle \pm 0.005}$ [3/3]
        & $0.292_{\scriptscriptstyle \pm 0.017}$ [3/3]
        & $0.297_{\scriptscriptstyle \pm 0.015}$ [3/3]
        & $0.290_{\scriptscriptstyle \pm 0.014}$ [3/3] \\

        MNLI & Accuracy ($\uparrow$)
        & $\infty$ [0/3]
        & $\infty$ [0/3]
        & $0.835_{\scriptscriptstyle \pm 0.002}$ [2/3]
        & $0.835_{\scriptscriptstyle \pm 0.003}$ [3/3] \\

        Language Modeling & Val. loss ($\downarrow$)
        & $4.931_{\scriptscriptstyle \pm 0.000}$ [1/3]
        & $4.923_{\scriptscriptstyle \pm 0.000}$ [1/3]
        & $\infty$ [0/3]
        & $4.927_{\scriptscriptstyle \pm 0.000}$ [1/3] \\

        Breakout & Avg. score ($\uparrow$)
        & $\infty$ [0/3]
        & $36.631_{\scriptscriptstyle \pm 3.010}$ [2/3]
        & $48.135_{\scriptscriptstyle \pm 4.396}$ [3/3]
        & $50.138_{\scriptscriptstyle \pm 12.057}$ [3/3] \\

        Mountain Car & Avg. reward ($\uparrow$)
        & $48.717_{\scriptscriptstyle \pm 0.000}$ [1/3]
        & $46.040_{\scriptscriptstyle \pm 10.138}$ [2/3]
        & $29.318_{\scriptscriptstyle \pm 19.455}$ [3/3]
        & $45.592_{\scriptscriptstyle \pm 3.126}$ [2/3] \\

        Meta Maze & Avg. return ($\uparrow$)
        & $24.981_{\scriptscriptstyle \pm 0.000}$ [1/3]
        & $\infty$ [0/3]
        & $8.091_{\scriptscriptstyle \pm 0.282}$ [2/3]
        & $21.732_{\scriptscriptstyle \pm 2.570}$ [3/3] \\

        3-SAT & Time (s) ($\downarrow$)
        & $13.665_{\scriptscriptstyle \pm 0.294}$ [3/3]
        & $11.973_{\scriptscriptstyle \pm 0.480}$ [3/3]
        & $11.621_{\scriptscriptstyle \pm 0.049}$ [3/3]
        & $12.740_{\scriptscriptstyle \pm 0.346}$ [3/3] \\

        \bottomrule
    \end{NiceTabular}
    \end{adjustbox}
    \vspace{3mm}
    \caption{\textbf{Per-task results with an open-weight trajectory teacher.} Values are mean and standard deviation over evaluation seeds 42, 123, and 456; brackets report valid runs. A dash denotes no valid solution. ML-AR students are trained on trajectories generated by Qwen3-32B over the fixed verified task pool.}
    \label{tab:open_weight_teacher_per_task}
\end{table*}

% \input{tables/mlgym_bench}

% \begin{figure}
%     \centering
%     \includegraphics[width=1\linewidth]{figures/diversity_exp.png}
%     \caption{\textbf{Demonstration of Research Topic Diversity.} Left: Overview of the diversity of topics generated during topic sampling. Right: Comparison of word distributions between all tasks and successful tasks.}
%     \label{fig:placeholder}
% \end{figure}

\subsection{Details of Ablation Study Results}
\label{appendix:ablation_details}
In this section, we provide detailed results of ablation studies across all tasks in the MLGym benchmark.
\begin{table*}[!h]
    \centering
    \begin{adjustbox}{width=\textwidth}
    \begin{NiceTabular}{llccc}
        \toprule
        Task & Metric & Z-score 0.1 & Z-score 0.5 & Z-score 1 \\
        \midrule
        CIFAR-10 & Accuracy ($\uparrow$) & 1.0000 & 1.0000 & 0.1426 \\
        Battle of Sexes & Average Reward ($\uparrow$) & 1.4418 & 2.0000 & 2.0000 \\
        Prisoners Dilemma & Average Reward ($\uparrow$) & 5.0000 & 2.6312 & 2.6284 \\
        Blotto & Average Reward ($\uparrow$) & 0.2458 & 0.2502 & 0.0790 \\
        House Price Prediction & $\text{R}^2$ Score ($\uparrow$) & 0.9106 & 0.9246 & 0.9200 \\
        Fashion MNIST & Accuracy ($\uparrow$) & 1.0000 & 0.9277 & 1.0000 \\
        MS-COCO & BLEU Score ($\uparrow$) & 0.9000 & 0.2806 & 0.2806 \\
        MNLI & Validation Accuracy ($\uparrow$) & 46.1233 & 63.2705 & 50.4432 \\
        Language Modeling & Validation Loss ($\downarrow$) & 4.9170 & 4.9187 & 4.9159 \\
        Breakout & Average Score ($\uparrow$) & 70.7970 & 56.8625 & 43.8539 \\
        Mountain Car Continuous & Average Reward ($\uparrow$) & 56.3959 & 49.4486 & 74.8382 \\
        Meta Maze & Average Return ($\uparrow$) & 48.6188 & 20.4422 & 20.1750 \\
        3-SAT Heuristic & Wall-Clock Time (s) ($\downarrow$) & 17.3700 & 19.4300 & 18.3060 \\
        \bottomrule
    \end{NiceTabular}
    \end{adjustbox}
    \caption{Comparison across different Z-score filtering thresholds. Arrows indicate whether higher ($\uparrow$) or lower ($\downarrow$) values are better.}
    \label{tab:zscore_comparison}
\end{table*}

\begin{table*}[!h]
    \centering
    \begin{adjustbox}{width=\textwidth}
    \begin{NiceTabular}{llccc}
        \toprule
        Task & Metric & No Filter & Z-score & Successful Experience \\
        \midrule
        CIFAR-10 & Accuracy ($\uparrow$) & $\infty$ & 0.1426 & 1.0000 \\
        Battle of Sexes & Average Reward ($\uparrow$) & 2.0000 & 2.0000 & 1.4453 \\
        Prisoners Dilemma & Average Reward ($\uparrow$) & 2.6283 & 2.6284 & 2.6324 \\
        Blotto & Average Reward ($\uparrow$) & 1.0000 & 0.0790 & 0.8631 \\
        House Price Prediction & $\text{R}^2$ Score ($\uparrow$) & 0.9246 & 0.9200 & 0.9274 \\
        Fashion MNIST & Accuracy ($\uparrow$) & 0.1000 & 1.0000 & 1.0000 \\
        MS-COCO & BLEU Score ($\uparrow$) & 0.2949 & 0.2806 & 0.3169 \\
        MNLI & Validation Accuracy ($\uparrow$) & 35.3337 & 50.4432 & 82.3943 \\
        Language Modeling & Validation Loss ($\downarrow$) & 4.9234 & 4.9159 & 4.9175 \\
        Breakout & Average Score ($\uparrow$) & 64.7484 & 43.8539 & 46.8381 \\
        Mountain Car Continuous & Average Reward ($\uparrow$) & 52.3536 & 74.8382 & 49.4486 \\
        Meta Maze & Average Return ($\uparrow$) & 22.3875 & 20.1750 & 23.8203 \\
        3-SAT Heuristic & Wall-Clock Time (s) ($\downarrow$) & 18.3470 & 18.3060 & 18.2600 \\
        \bottomrule
    \end{NiceTabular}
    \end{adjustbox}
    \caption{Comparison across filtering strategies. Arrows indicate whether higher ($\uparrow$) or lower ($\downarrow$) values are better.}
    \label{tab:ba_raw_updated}
\end{table*}

\begin{table*}[!h]
    \centering
    \begin{adjustbox}{width=\textwidth}
    \begin{NiceTabular}{llcc}
        \toprule
        Task & Metric & 32k & 64k \\
        \midrule
        CIFAR-10 & Accuracy ($\uparrow$) & 0.4977 & 0.4977 \\
        Battle of Sexes & Average Reward ($\uparrow$) & 1.8032 & 1.4405 \\
        Prisoners Dilemma & Average Reward ($\uparrow$) & 3.3925 & 2.3817 \\
        Blotto & Average Reward ($\uparrow$) & 1.0000 & 1.0000 \\
        House Price Prediction & $\text{R}^2$ Score ($\uparrow$) & 0.8800 & 0.8800 \\
        Fashion MNIST & Accuracy ($\uparrow$) & 0.8480 & 0.8480 \\
        MS-COCO & BLEU Score ($\uparrow$) & 0.2806 & 0.2806 \\
        MNLI & Validation Accuracy ($\uparrow$) & 54.2129 & 55.1707 \\
        Language Modeling & Validation Loss ($\downarrow$) & 4.9191 & 4.9222 \\
        Breakout & Average Score ($\uparrow$) & 46.8381 & 46.8381 \\
        Mountain Car Continuous & Average Reward ($\uparrow$) & 49.4486 & 49.4486 \\
        Meta Maze & Average Return ($\uparrow$) & 23.6078 & 23.8281 \\
        3-SAT Heuristic & Wall-Clock Time (s) ($\downarrow$) & 17.9490 & 17.8090 \\
        \bottomrule
    \end{NiceTabular}
    \end{adjustbox}
    \caption{Comparison across different context lengths. Arrows indicate whether higher ($\uparrow$) or lower ($\downarrow$) values are better.}
    \label{tab:token_len_comparison}
\end{table*}

\begin{table*}[!h]
    \centering
    \begin{adjustbox}{width=\textwidth}
    \begin{NiceTabular}{llcc}
        \toprule
        Task & Metric & Qwen2.5-3B-Instruct & ML-AR-3B \\
        \midrule
        CIFAR-10 & Accuracy ($\uparrow$) & $\infty$ & 0.0996 \\
        Battle of Sexes & Average Reward ($\uparrow$) & 1.0304 & 1.4378 \\
        Prisoners Dilemma & Average Reward ($\uparrow$) & 2.3883 & 2.6225 \\
        Blotto & Average Reward ($\uparrow$) & -0.2430 & 1.0000 \\
        House Price Prediction & $\text{R}^2$ Score ($\uparrow$) & -4.1698 & 0.8800 \\
        Fashion MNIST & Accuracy ($\uparrow$) & $\infty$ & $\infty$ \\
        MS-COCO & BLEU Score ($\uparrow$) & $\infty$ & 0.2809 \\
        MNLI & Validation Accuracy ($\uparrow$) & $\infty$ & $\infty$ \\
        Language Modeling & Validation Loss ($\downarrow$) & $\infty$ & $\infty$ \\
        Breakout & Average Score ($\uparrow$) & 28.1272 & $\infty$ \\
        Mountain Car Continuous & Average Reward ($\uparrow$) & $\infty$ & $\infty$ \\
        Meta Maze & Average Return ($\uparrow$) & $\infty$ & $\infty$ \\
        3-SAT Heuristic & Wall-Clock Time (s) ($\downarrow$) & 18.0780 & 18.2730 \\
        \bottomrule
    \end{NiceTabular}
    \end{adjustbox}
    \caption{Comparison between Qwen2.5-3B-Instruct and ML-AR-3B. Arrows indicate whether higher ($\uparrow$) or lower ($\downarrow$) values are better.}
    \label{tab:model_family_comparison}
\end{table*}

\subsection{Broader impact}
\label{sec:appendix-broader-impacts}
The development of autonomous machine learning agents presents significant dual-use characteristics, carrying both profound positive potential and notable risks of misuse. 

\paragraph{Positive Societal Impacts.} 
Frameworks like ML-AutoResearch (ML-AR) have the potential to democratize scientific discovery by automating the highly technical, engineering-heavy, and time-consuming aspects of ML research. By lowering the barrier to entry for developing and optimizing machine learning models, such systems can drastically reduce the cost of research and development. This acceleration allows human researchers to shift their focus away from tedious debugging and hyperparameter tuning, and toward high-level conceptual design, safety alignment, and solving complex societal challenges.

\paragraph{Negative Misuse Risks.} 
Conversely, these advanced capabilities carry inherent risks. An agent capable of autonomously writing, debugging, and optimizing ML code could be repurposed by malicious actors to rapidly develop harmful AI systems. This includes accelerating the creation of advanced surveillance tools, generating highly convincing deepfakes, or automating the discovery of vulnerabilities for cyberattacks. Furthermore, the autonomous nature of these agents means that, without proper oversight, they might inadvertently optimize for proxy metrics that introduce severe biases, unfairness, or security flaws into deployed models.

\paragraph{Mitigations and Evaluation Context.} 
To mitigate these risks during the development and evaluation phases, our work strictly confines agent execution to controlled, sandboxed environments such as the MLGym. This environment isolates the agents from the open internet, restricting them to well-defined datasets and scoped computational resources, thereby preventing the execution of arbitrary, unverified code on host machines. As autonomous ML research systems continue to scale, we strongly advocate for the continued use of secure sandboxing, rigorous alignment checks, and human-in-the-loop oversight before such agents are deployed in unconstrained, real-world settings.

\subsection{Licenses and terms of use}
\label{sec:appendix-license-audit}

We document licenses and terms for third-party assets used in this work in the
following inventory.

\begin{table}[h]
\centering
\small
\begin{tabular}{p{0.22\linewidth}p{0.24\linewidth}p{0.30\linewidth}p{0.16\linewidth}}
\toprule
Asset class & Example assets in this work & Licenses/terms used & Notes \\
\midrule
Datasets & Hugging Face datasets used for synthetic task generation and evaluation & Apache-2.0, MIT, CC0-1.0, CC-BY-4.0, CC-BY-SA-4.0, CC-BY-SA-3.0, CC-BY-SA-3.0+GFDL, BSD/BSD-3-Clause, CC-BY-NC-SA-4.0, CC-BY-NC-ND-4.0, AFL-3.0, ODC-BY, C-UDA & Per-dataset terms are respected and attributed \\
Open models & Qwen family student models & Qwen model licenses/terms from the corresponding model cards & Version-specific model terms apply \\
API models & GPT teacher/evaluator APIs & OpenAI API Terms of Use and associated service policies & API use only; no model weight redistribution \\
Benchmarks/codebases & MLGym, MLAgentBench, AutoResearch benchmark code & Repository-specific OSS licenses (including MIT/Apache-2.0/BSD-style where applicable) & We follow each repository's LICENSE file \\
Agent framework/tools & SWE-agent and related tooling & Repository-specific OSS licenses (including MIT/Apache-2.0/BSD-style where applicable) & We follow each repository's LICENSE file \\
\bottomrule
\end{tabular}
\caption{Licenses and terms of use for external assets used in this work.}
\label{tab:dataset_license_audit}
\end{table}

All external assets are used under their respective licenses or terms of service.

\subsection{Reproducibility and training details}
\label{sec:appendix-repro}

We provide implementation-level details used for SFT runs in our internal training
pipeline, including data construction, model optimization, filtering, and cluster
compute settings.

\begin{table}[h]
\centering
\small
\begin{tabular}{p{0.24\linewidth}p{0.68\linewidth}}
\toprule
Item & Setting \\
\midrule
Base models & Qwen3-4B, Qwen3-8B, Qwen3-14B \\
Tokenizer & Matching Qwen tokenizer for each base model \\
Optimizer & AdamW \\
Learning rate & $1.0\times10^{-5}$ (4B), $5.0\times10^{-6}$ (8B), $2.0\times10^{-5}$ (14B) on H100 configs \\
LR schedule & WarmupDecayLR with linear warmup; warmup steps 100 (4B/8B), 200 (14B) \\
Weight decay & $1.0\times10^{-4}$ (4B/8B), $1.0\times10^{-1}$ (14B config) \\
Global batch size & 64 (4B/8B), 32 (14B) \\
Micro-batch size/GPU & 2 (4B/8B H100), 1 (14B H100) \\
Training length & 1500 max steps (H100 configs) \\
Precision & BF16 enabled \\
Random seed & 42 \\
Sequence window & 64k in current data build (\texttt{max\_seq\_len=65536}); trajectories above \texttt{max\_seq\_len+8192} are filtered out \\
Trajectory selection & Per-task metric-aware ranking; optional z-score filtering (\texttt{MLGYM\_DROP\_Z\_THRESHOLD}, default 1.0) and top-percentile retention (\texttt{MLGYM\_KEEP\_TOP\_PERCENTILE}) \\
\bottomrule
\end{tabular}
\caption{Training and data-construction settings used by the SFT pipeline.}
\label{tab:repro_details}
\end{table}

\subsection{Prompts used in the task generation pipeline}
\label{sec:appendix-prompts}

This appendix lists the core, non-redundant prompt texts used in the data generation pipeline.

\subsubsection{Topic sampling prompt}
\begin{tcolorbox}[breakable, colback=gray!5, colframe=gray!40, boxrule=0.5pt, sharp corners]
\begin{lstlisting}[basicstyle=\ttfamily\small, breaklines=true, columns=fullflexible]
You are an expert in machine learning research. Generate a list of 20 diverse and interesting machine learning research topics. Each topic should be a short phrase or sentence, suitable for use as a research challenge or task. Do not repeat topics from previous examples. Return the topics as a JSON array of strings.

Your output:
\end{lstlisting}
\end{tcolorbox}

\subsection{Task proposal and dataset validation prompts}
\paragraph{Task proposal prompts}
\begin{tcolorbox}[breakable, colback=gray!5, colframe=gray!40, boxrule=0.5pt, sharp corners]
\begin{lstlisting}[basicstyle=\ttfamily\small, breaklines=true, columns=fullflexible]
You are an expert ML researcher create a training task for a junior researcher. Given a topic, generate a JSON object describing a machine learning task for that topic. The JSON must include:
- topic: the original topic
- metric: a suitable evaluation metric for the task
- description: a detailed description of the ML task
- dataset: a dataset name that can be matched to a huggingface dataset OR a simple search query for a public huggingface dataset (e.g., 'cifar10', 'imdb', 'tiny imagenet'). Omit this field if the task does not require a dataset.

You have access to a tool that can search the HuggingFace datasets API to find suitable datasets based on your query. Please make sure the dataset exists on HuggingFace.

Here are some examples:

Topic: Image Classification
Output:
{"topic": "Image Classification", "metric": "Accuracy", "description": "Classify images into categories using a standard image classification dataset.", "dataset": "cifar10"}

Topic: Sentiment Analysis
Output:
{"topic": "Sentiment Analysis", "metric": "Accuracy", "description": "Predict the sentiment (positive/negative) of movie reviews.", "dataset": "imdb"}

Topic: Named Entity Recognition
Output:
{"topic": "Named Entity Recognition", "metric": "F1-score", "description": "Identify named entities in text using a standard NER dataset.", "dataset": "conll2003"}

Topic: Text Summarization
Output:
{"topic": "Text Summarization", "metric": "ROUGE score", "description": "Generate concise summaries of news articles.", "dataset": "cnn_dailymail"}

Topic: Machine Translation
Output:
{"topic": "Machine Translation", "metric": "BLEU", "description": "Translate sentences from English to German.", "dataset": "wmt14"}

Topic: Speech Command Recognition
Output:
{"topic": "Speech Command Recognition", "metric": "Accuracy", "description": "Classify spoken commands from audio clips.", "dataset": "google speech commands"}

Topic: Human Activity Recognition
Output:
{"topic": "Human Activity Recognition", "metric": "Accuracy", "description": "Classify human activities from wearable sensor data.", "dataset": "UCI HAR"}

Topic: {{topic}}
Output:
\end{lstlisting}
\end{tcolorbox}

\paragraph{Dataset validation prompt}
\begin{tcolorbox}[breakable, colback=gray!5, colframe=gray!40, boxrule=0.5pt, sharp corners]
\begin{lstlisting}[basicstyle=\ttfamily\small, breaklines=true, columns=fullflexible]
You may call the dataset search tool to validate or refine the dataset choice before producing the final JSON.
Rules:
- If you are unsure about the dataset name, call the tool with a short query.
- When confident, output ONLY the final JSON object (no surrounding prose) with required keys.
- Keys: topic, metric, description, optional dataset (string). If dataset provided, ensure it plausibly exists on HuggingFace.
- Avoid re-calling the tool if current results already contain a suitable dataset.
- You can modify the topic slightly to fit the available datasets.
\end{lstlisting}
\end{tcolorbox}

\paragraph{Dataset Search Tool-Result Follow-Up Prompt}
\begin{tcolorbox}[breakable, colback=gray!5, colframe=gray!40, boxrule=0.5pt, sharp corners]
\begin{lstlisting}[basicstyle=\ttfamily\small, breaklines=true, columns=fullflexible]
Search results for query '{query}': {json.dumps(results, ensure_ascii=False)}
Select one dataset id (or refine by calling the tool again) and output final JSON when ready.
\end{lstlisting}
\end{tcolorbox}

\paragraph{JSON-Missing Nudge Prompt}
\begin{tcolorbox}[breakable, colback=gray!5, colframe=gray!40, boxrule=0.5pt, sharp corners]
\begin{lstlisting}[basicstyle=\ttfamily\small, breaklines=true, columns=fullflexible]
I did not receive a valid JSON. Please either call the search tool or output the final JSON object now.
\end{lstlisting}
\end{tcolorbox}

\subsection{Task files generation prompt}
\paragraph{Task files stage 1: config generation}
\begin{tcolorbox}[breakable, colback=gray!5, colframe=gray!40, boxrule=0.5pt, sharp corners]
\begin{lstlisting}[basicstyle=\ttfamily\small, breaklines=true, columns=fullflexible]
Your objective is to create YAML config files for a machine learning task. You are given JSON input that describes the topic as well as the dataset you are working with. The first file is a task configuration file that describes the task, dataset, and submission format. The other files (usually one but can be multiple) are dataset configuration files that describe the datasets used in the task. Be creative and generate a task that is interesting and challenging for the agent to solve.

IMPORTANT OUTPUT FORMAT REQUIREMENT:
Return every file using markdown code blocks ONLY (no sentinel markers, no extra text) exactly like:
```lang
# relative/path/to/file
<file contents>
```

- The task is executed in a linux environment with Python 3.10 and the following packages preinstalled (generic_conda_requirements.txt):
 generic_conda_requirements.txt (preinstalled packages):
  numpy
  pandas
  scipy
  torch
  scikit-learn
  tqdm
  datasets
  gymnasium
  transformers[torch]
  datasets
  matplotlib
  torchvision

Here is the format for the input JSON.
```json
# input.json
{
  "topic": ...,
  "metric": ...,
  "description": ...,
  "dataset": {
    "id": ...,
    "features": [...],
    "examples": [...]
  }
}
```

Here is the format for the config files (showing expected file outputs using markdown code blocks):
```yaml
# tasks/task_id.yaml
id: # task id, will be used as the config file name as well. Recommend using snake_case.
name: # task name
description: # Description of the task that includes the task objective submission format requirements. You must include the string "{dataset_docs}", which reference the description in the dataset config file.
dataset_configs: # zero or more data config files described below.
task_entrypoint: # one of four values: CSVSubmissionTasks, ModelSubmissionTasks, LMSubmissionTasks, PythonSubmissionTasks
training_timeout: # timeout in seconds, make a good effort estimating the time it takes to train the model on our hardware (NVIDIA RTX A6000 GPU, 8 CPU cores).
use_generic_conda: # true if the task does not require any packages other than the ones listed in generic_conda_requirements.txt, false otherwise
requirements_path: # path to requirements.txt if use_generic_conda is false, otherwise leave this empty
starter_code: [] # Leave this as an empty list, it will be filled in later
baseline_paths: [] # Leave this empty, it will be filled in later
baseline_scores: [] # Leave this as an empty list, it will be filled in later
evaluation_paths: [] # Leave this empty, it will be filled in later
evaluation_read_only: # Whether the evaluation script should be read-only to the agent
memory_path: memory.json # This value is fixed
```

```yaml
# datasets/dataset_name.yaml
data_path: # Path to the dataset, IMPORTANT: This must be a valid public huggingface dataset, for example "uoft-cs/cifar10" or "ILSVRC/imagenet-1k". You can find the dataset ID in the provided JSON input. Do not use a placeholder.
description: # detailed description of the dataset, including features, content, format, number of classes and samples, and any other relevant information. Give concrete example rows of the dataset.
is_local: # should always be false
name: # dataset name
```

You can only use one of four values for `task_entrypoint` outlined below.

## Quick Decision Flow

* Agent outputs a CSV predictions file? -> CSVSubmissionTasks
* Agent submits a model/checkpoint + YAML config? -> ModelSubmissionTasks
* Language-model training/eval that should run with torchrun on GPUs? -> LMSubmissionTasks
* Deliverable is Python code you evaluate directly? -> PythonSubmissionTasks

Set this with task_entrypoint in your TaskConfig.

---
1) CSVSubmissionTasks
Use when: The submission is a CSV of predictions (Kaggle-style).
Submission expected: submission.csv in the task workspace root.
Evaluation call (first path in evaluation_paths):
 python <eval_script> --submission_file <path/to/submission.csv>
Eval output format: Entire stdout must be a valid JSON object.
Baseline: If baseline_paths is set, runs the first baseline script, then evaluate().
Config snippet:
 task_entrypoint: CSVSubmissionTasks

2) ModelSubmissionTasks
Use when: The agent submits a model artifact or config (e.g., checkpoints + a YAML config), not a CSV.
Submission expected: The first *.yaml file found under the workspace.
Evaluation call:
 python <eval_script> --config_fname <path/to/config.yaml>
Eval output format: Stdout must contain at least one line starting with { that is valid JSON.
Baseline: Same approach.
Config snippet:
 task_entrypoint: ModelSubmissionTasks

3) LMSubmissionTasks
Use when: Language-model tasks that should run distributed via torchrun.
Evaluation call:
 torchrun --nproc_per_node=<detected_gpus> --standalone <eval_script>
Eval output format: First line starting with { that parses as JSON.
Baseline: Also with torchrun.
Config snippet:
 task_entrypoint: LMSubmissionTasks

4) PythonSubmissionTasks
Use when: The agent writes Python code that evaluator imports/executes directly.
Submission expected: target.py file.
Evaluation call:
 python <eval_script>
Eval output format: Entire stdout JSON object.
Config snippet:
 task_entrypoint: PythonSubmissionTasks
---

Important tips:
1. Make the task and dataset description as detailed as possible: task objective, dataset format, data examples, submission format, metrics, constraints. Be very informative because the agent rely on this information.
2. Include full examples of data rows in the dataset config description.
3. The "id" field of the task config must be exactly same as the filename without the .yaml extension.
4. In the description, always escape curly braces with double braces {{ and }}, except for {dataset_docs}.
5. You may see a dataset field in the input JSON, use that as the data_path in the dataset config. Use the dataset ID from the input JSON exactly as the dataset name, it is a verified public huggingface dataset.
6. Use the dataset information provided to you (if any), give detailed information about the features, content, and example rows of the dataset (if any).
7. If you are not prompted with a dataset, optionally choose a valid public huggingface dataset. DO NOT emit a placeholder or other invalid names of the dataset.
8. Follow the task description provided in the input.
9. Output files in order: one task config, then dataset config files. Only one task config file.

Example description:
{{example_input_1}}

Example output:
{{example_1}}

Example description:
{{example_input_2}}

Example output:
{{example_2}}

Your description:
{{task_description}}

Think step by step and plan out the task before writing the output files.
Your output:
\end{lstlisting}
\end{tcolorbox}

\paragraph{Task files stage 2: starter code generation}
\begin{tcolorbox}[breakable, colback=gray!5, colframe=gray!40, boxrule=0.5pt, sharp corners]
\begin{lstlisting}[basicstyle=\ttfamily\small, breaklines=true, columns=fullflexible]
You are tasked to create a ML training task for a autonomous machine learning research agent according to given config files.
The agent is an autonomous Machine Learning Researcher operating in a specialized command-line environment.
In a turn-based interaction, the agent provides a "discussion" of its plan, followed by a single shell command.
The agent can navigate the file system, and read and write files using special commands, but must handle code indentation manually.
The agent is provided with baseline code for an ML task and its goal is to improve the model's performance and submit the final solution.

The input task config file look like this:
```yaml
# tasks/task_id.yaml
id: # task id, will be used as the config file name as well. Recommend using snake_case.
name: # task name
description: # Description of the task that includes the task objective submission format requirements. You must include the string "{dataset_docs}", which reference the description in the dataset config file.
dataset_configs: # zero or more data config files described below.
task_entrypoint: # one of four values: CSVSubmissionTasks, ModelSubmissionTasks, LMSubmissionTasks, PythonSubmissionTasks
training_timeout: # timeout in seconds, make a good effort estimating the time it takes to train the model on our hardware (NVIDIA RTX A6000 GPU, 8 CPU cores).
use_generic_conda: # true if the task does not require any packages other than the ones listed in generic_conda_requirements.txt, false otherwise
requirements_path: # path to requirements.txt if use_generic_conda is false, otherwise leave this empty
starter_code: [] # Fill this in after creating the task files
baseline_paths: [] # You need to identify the baseline file and fill this in
baseline_scores: [] # Lease this empty, it will be filled in later by running your evaluation
evaluation_paths: [] # You need to identify the evaluation file and fill this in
evaluation_read_only: # Whether the evaluation script should be read-only to the agent
memory_path: memory.json # This value is fixed
```

The input dataset configs are optional, and look like this:
```yaml
# datasets/dataset_name.yaml
data_path: # Path to the dataset, IMPORTANT: This must be a valid public huggingface dataset, for example "uoft-cs/cifar10" or "ILSVRC/imagenet-1k". Do not use a placeholder.
description: # detailed description of the dataset, including content, format, number of classes and samples, and any other relevant information
is_local: # should always be false
name: # dataset name
```

IMPORTANT OUTPUT FORMAT REQUIREMENT:
Return every file using markdown code blocks ONLY exactly like:
```lang
# relative/path/to/file
<file contents>
```

Output ALL created files this way.

The task is executed in a linux environment with Python 3.10 and the following packages preinstalled (generic_conda_requirements.txt):
 generic_conda_requirements.txt (preinstalled packages):
  numpy
  pandas
  scipy
  torch
  scikit-learn
  tqdm
  datasets
  gymnasium
  transformers[torch]
  datasets
  matplotlib
  torchvision
If you decides to use any other packages, you MUST include a file named requirements.txt in the output, and set the use_generic_conda field to false in the task config.


Follow the following instructions:
- The input gives you a task config + dataset config(s). You must produce runnable starter code.
- Leave the starter_code field empty, I will help you fill it.
- Provide exactly one baseline file path in baseline_paths; baseline must be directly runnable without any command line arguments and generate a valid submission artifact.
- Provide exactly one evaluation file path in evaluation_paths; IMPORTANT: evaluation must output a JSON with a single field and numeric score to stdout.
- Do not fill baseline_scores, it will be filled in later by running the baseline and evaluation.
- You MAY create auxiliary scripts (data utils, model, etc.)
- Evaluation file MUST print a single valid JSON object with string keys and float values (only once) for metrics.
- Task must run in under 30 minutes on a NVIDIA RTX A6000 GPU and 8 CPU cores.
- If you set use_generic_conda: true, then use only the preinstalled packages. If you set use_generic_conda: false you add requirements.txt, the path to requirements.txt MUST be set in the task config.
- Choose dataset usage consistent with provided dataset config(s) and task type. Remember you MUST use a REAL and VALID public huggingface dataset. Your task may not need a dataset (e.g. game theory).
- Keep the baseline simple, leave room for the agent to improve it.
- If the user runs into errors validating the task, you can change the task config to fix the issue.


When you create the evaluation script, if will be ran with different commands based on the value of the task_entrpoint field. The output must print a valid JSON object. Your evaluation file must respect the evaluation call format for the task class.
---
1) CSVSubmissionTasks
Use when: The submission is a CSV of predictions (Kaggle-style).
Submission expected: submission.csv in the task workspace root.
Evaluation call (first path in evaluation_paths):
 python <eval_script> --submission_file <path/to/submission.csv>

2) ModelSubmissionTasks
Use when: The agent submits a model artifact or config (e.g., checkpoints + a YAML config), not a CSV.
Submission expected: The first *.yaml file found under the workspace.
Evaluation call:
 python <eval_script> --config_fname <path/to/config.yaml>

3) LMSubmissionTasks
Use when: Language-model tasks that should run distributed via torchrun.
Evaluation call:
 torchrun --nproc_per_node=<detected_gpus> --standalone <eval_script>
Eval output format: First line starting with { that parses as JSON.

4) PythonSubmissionTasks
Use when: The agent writes Python code that evaluator imports/executes directly.
Submission expected: target.py file.
Evaluation call:
 python <eval_script>
---

Coding tips:
- Use multiprocessing / DataLoader workers for speed. You have 8 CPU cores.
- Use GPUs for training and evaluation and anything else that makes sense, assume it is always available. You have a NVIDIA RTX A6000 GPU.
- Use deterministic seeds where relevant.
- Use the right indentation.
- Import the dataset using the datasets library load_dataset function.
- Do not import scripts that you have not written.
- Do not import packages that are not in the generic_conda_requirements.txt or requirements.txt.
- For your ease, strongly prefer a flat directory structure, and create subfolders only if necessary.
- Use the dataset_path variable in the dataset config given to you. It is verified to be a valid public huggingface dataset.
- Try to use as flexible as possible package requirements, e.g. "torch>=2.0.0" instead of "torch==2.0.0" in requirements.txt, since you may have outdated knowledge.

Here is an example:
{{example_input_1}}
Example output:
{{example_output_1}}

Here is another example:
{{example_input_2}}
Example output:
{{example_output_2}}

Here is your task config:
{{task_config}}

Think step by step and plan out the task before writing the code files.
Your output:
\end{lstlisting}
\end{tcolorbox}

\paragraph{Error-Recovery Retry Prompt}
\begin{tcolorbox}[breakable, colback=gray!5, colframe=gray!40, boxrule=0.5pt, sharp corners]
\begin{lstlisting}[basicstyle=\ttfamily\small, breaklines=true, columns=fullflexible]
Error encountered: {self.stage_1_err}

Please try again. Return the revised output in whole.
\end{lstlisting}
\end{tcolorbox}

\subsection{Example synthetic task}
We show a random example among our generated tasks. The task includes \begin{enumerate}
    \item Task description \texttt{hotpotqa\_join\_facts\_qa.yaml}
    \item Dataset description \texttt{hotpotqa\_hotpot\_qa.yaml}
    \item Starting implementation \texttt{baseline.py}
    \item Evaluation code \texttt{evaluate.py}
\end{enumerate}

\paragraph{hotpotqa\_hotpot\_qa.yaml}hotpotqa\_join\_facts\_qa.yaml
\begin{tcolorbox}[breakable, colback=gray!5, colframe=gray!40, boxrule=0.5pt, sharp corners]
    \begin{lstlisting}[basicstyle=\ttfamily\small, breaklines=true, columns=fullflexible]
data_path: hotpotqa/hotpot_qa
description: "HotpotQA is a large-scale multi-hop question answering dataset featuring\
  \ questions that require reasoning across multiple documents. This configuration\
  \ targets the distractor setting, where each example provides 10 candidate paragraphs\
  \ (titles and sentence lists), of which only a subset contains the gold supporting\
  \ sentences needed to answer the question.\nFeatures: - id (string): Unique identifier\
  \ for the example, e.g., \"5a7a06935542990198eaf050\". - question (string): Natural\
  \ language question requiring multi-hop reasoning. - answer (string): Gold answer\
  \ text (can be \"yes\"/\"no\" or a short span). - type (string): Question type,\
  \ e.g., \"comparison\", \"bridge\". - level (string): Difficulty level, e.g., \"\
  easy\", \"medium\", \"hard\". - supporting_facts (struct of lists):\n    - title\
  \ (list[string]): Titles of the documents containing supporting sentences.\n   \
  \ - sent_id (list[int32]): 0-based indices of the supporting sentences in the corresponding\
  \ documents.\n  The k-th title aligns with the k-th sent_id to form a pair (title[k],\
  \ sent_id[k]).\n- context (struct):\n    - title (list[string]): Titles of the 10\
  \ candidate documents.\n    - sentences (list[list[string]]): For each document,\
  \ a list of its sentence strings, aligned by index with context.title.\n\nTypical\
  \ splits: - train: ~90k-113k examples (depending on release/version). - validation/dev:\
  \ ~7k-8k examples. - test: may be available without supporting facts/answers in\
  \ certain releases. For this task, use train and validation/dev.\nData format details:\
  \ - Sentence indices in supporting_facts.sent_id are 0-based and reference the sentence\
  \ array of the document whose title matches supporting_facts.title. - The distractor\
  \ setting includes 10 documents (context.title length == 10); each has a variable\
  \ number of sentences. - Answer normalization for EM/F1 is performed during evaluation\
  \ (lowercasing, removing punctuation and articles).\nConcrete examples: - Example\
  \ 1:\n  id: \"5a7a06935542990198eaf050\"\n  question: \"Which magazine was started\
  \ first Arthur's Magazine or First for Women?\"\n  answer: \"Arthur's Magazine\"\
  \n  type: \"comparison\"\n  level: \"medium\"\n  supporting_facts:\n    title: [\"\
  Arthur's Magazine\", \"First for Women\"]\n    sent_id: [0, 0]\n  context:\n   \
  \ title: [\n      \"Radio City (Indian radio station)\", \"History of Albanian football\"\
  , \"Echosmith\",\n      \"Women's colleges in the Southern United States\", \"First\
  \ Arthur County Courthouse and Jail\",\n      \"Arthur's Magazine\", \"2014-15 Ukrainian\
  \ Hockey Championship\", \"First for Women\",\n      \"Freeway Complex Fire\", \"\
  William Rast\"\n    ]\n    sentences: [\n      [\n        \"Radio City is India's\
  \ first private FM radio station and was started on 3 July 2001.\",\n        \"\
  It broadcasts on 91.1 (earlier 91.0 in most cities) megahertz from Mumbai (where\
  \ it was started in 2004), Bengaluru (started first in 2001), Lucknow and New Delhi\
  \ (since 2003).\",\n        \"It plays Hindi, English and regional songs.\",\n \
  \       \"It was launched in Hyderabad in March 2006, in Chennai on 7 July 2006\
  \ and in Visakhapatnam October 2007.\",\n        \"Radio City recently forayed into\
  \ New Media in May 2008 with the launch of a music portal - PlanetRadiocity.com\
  \ that offers music related news, videos, songs, and other music-related features.\"\
  ,\n        \"The Radio station currently plays a mix of Hindi and Regional music.\"\
  ,\n        \"Abraham Thomas is the CEO of the company.\"\n      ],\n      ...\n\
  \      [\n        \"Arthur's Magazine (1844-1846) was an American literary periodical\
  \ published in Philadelphia in the 19th century.\",\n        \"Edited by T.S. Arthur,\
  \ it featured work by Edgar A. Poe, J.H. Ingraham, Sarah Josepha Hale, Thomas G.\
  \ Spear, and others.\",\n        \"In May 1846 it was merged into \\\"Godey's Lady's\
  \ Book\\\".\"\n      ],\n      ...\n      [\n        \"First for Women is a woman's\
  \ magazine published by Bauer Media Group in the USA.\",\n        \"The magazine\
  \ was started in 1989.\",\n        \"It is based in Englewood Cliffs, New Jersey.\"\
  ,\n        \"In 2011 the circulation of the magazine was 1,310,696 copies.\"\n \
  \     ],\n      ...\n    ]\n\n- Example 2:\n  id: \"5a879ab05542996e4f30887e\"\n\
  \  question: \"The Oberoi family is part of a hotel company that has a head office\
  \ in what city?\"\n  answer: \"Delhi\"\n  type: \"bridge\"\n  level: \"medium\"\n\
  \  supporting_facts:\n    title: [\"Oberoi family\", \"The Oberoi Group\"]\n   \
  \ sent_id: [0, 0]\n  context:\n    title: [\n      \"Ritz-Carlton Jakarta\", \"\
  Oberoi family\", \"Ishqbaaaz\", \"Hotel Tallcorn\", \"Mohan Singh Oberoi\",\n  \
  \    \"Hotel Bond\", \"The Oberoi Group\", \"Future Fibre Technologies\", \"289th\
  \ Military Police Company\",\n      \"Glennwanis Hotel\"\n    ]\n    sentences:\
  \ [\n      [\n        \"The Ritz-Carlton Jakarta is a hotel and skyscraper in Jakarta,\
  \ Indonesia and 14th Tallest building in Jakarta.\",\n        \"It is located in\
  \ city center of Jakarta, near Mega Kuningan, adjacent to the sister JW Marriott\
  \ Hotel.\",\n        \"It is operated by The Ritz-Carlton Hotel Company.\",\n  \
  \      \"The complex has two towers that comprises a hotel and the Airlangga Apartment\
  \ respectively.\",\n        \"The hotel was opened in 2005.\"\n      ],\n      [\n\
  \        \"The Oberoi family is an Indian family that is famous for its involvement\
  \ in hotels, namely through The Oberoi Group.\"\n      ],\n      ...,\n      [\n\
  \        \"The Oberoi Group is a hotel company with its head office in Delhi.\"\
  ,\n        \"Founded in 1934, the company owns and/or operates 30+ luxury hotels\
  \ and two river cruise ships in six countries, primarily under its Oberoi Hotels\
  \ & Resorts and Trident Hotels brands.\"\n      ],\n      ...\n    ]\n\n- Example\
  \ 3:\n  id: \"5a8d7341554299441c6b9fe5\"\n  question: \"Musician and satirist Allie\
  \ Goertz wrote a song about the \\\"The Simpsons\\\" character Milhouse, who Matt\
  \ Groening named after who?\"\n  answer: \"President Richard Nixon\"\n  type: \"\
  bridge\"\n  level: \"hard\"\n  supporting_facts:\n    title: [\"Allie Goertz\",\
  \ \"Allie Goertz\", \"Allie Goertz\", \"Milhouse Van Houten\"]\n    sent_id: [0,\
  \ 1, 2, 0]\n  context:\n    title: [\n      \"Lisa Simpson\", \"Marge Simpson\"\
  , \"Bart Simpson\", \"Allie Goertz\", \"Milhouse Van Houten\",\n      \"Los Angeles\
  \ Reader\", \"Homer Simpson\", \"List of The Simpsons video games\",\n      \"The\
  \ Simpsons: An Uncensored, Unauthorized History\", \"List of The Simpsons guest\
  \ stars\"\n    ]\n    sentences: [\n      [\n        \"Lisa Marie Simpson is a fictional\
  \ character in the animated television series \\\"The Simpsons\\\".\",\n       \
  \ \"She is the middle child and most intelligent of the Simpson family.\",\n   \
  \     \"Voiced by Yeardley Smith, Lisa first appeared on television in \\\"The Tracey\
  \ Ullman Show\\\" short \\\"Good Night\\\" on April 19, 1987.\",\n        \"Cartoonist\
  \ Matt Groening created and designed her while waiting to meet James L. Brooks.\"\
  ,\n        \"Groening had been invited to pitch a series of shorts based on his\
  \ comic \\\"Life in Hell\\\", but instead decided to create a new set of characters.\"\
  ,\n        \"He named the elder Simpson daughter after his younger sister Lisa Groening.\"\
  ,\n        \"After appearing on \\\"The Tracey Ullman Show\\\" for three years,\
  \ the Simpson family were moved to their own series on Fox, which debuted on December\
  \ 17, 1989.\"\n      ],\n      ...,\n      [\n        \"Allison Beth \\\"Allie\\\
  \" Goertz (born March 2, 1991) is an American musician.\",\n        \"Goertz is\
  \ known for her satirical songs based on various pop culture topics.\",\n      \
  \  \"Her videos are posted on YouTube under the name of Cossbysweater.\",\n    \
  \    \"Subjects of her songs have included the film \\\"The Room\\\", the character\
  \ Milhouse from the television show \\\"The Simpsons\\\", and the game Dungeons\
  \ & Dragons.\",\n        \"Her style has been compared to that of Bo Burnham.\"\
  ,\n        \"In December 2015, Goertz released a concept album based on the Adult\
  \ Swim series \\\"Rick and Morty\\\", \\\"Sad Dance Songs\\\", with the album's\
  \ cover emulating the animation and logo of the series.\",\n        \"The album\
  \ was made possible through Kickstarter.\",\n        \"She is co-host of Everything's\
  \ Coming Up Podcast, a Simpsons-focused podcast along with Julia Prescott.\"\n \
  \     ],\n      [\n        \"Milhouse Mussolini van Houten is a fictional character\
  \ featured in the animated television series \\\"The Simpsons\\\", voiced by Pamela\
  \ Hayden, and created by Matt Groening who named the character after President Richard\
  \ Nixon's middle name.\",\n        \"Later in the series, it is revealed that Milhouse's\
  \ middle name is \\\"Mussolini.\\\"\"\n      ],\n      ...\n    ]\n\nNotes: - Titles\
  \ must match exactly when referencing supporting facts. - sent_id values must be\
  \ valid indices into the corresponding document's sentence list. - The dataset is\
  \ English-only and licensed under CC BY-SA 4.0.\n"
is_local: false
name: hotpotqa/hotpot_qa
    \end{lstlisting}
\end{tcolorbox}

\paragraph{hotpotqa\_join\_facts\_qa.yaml}
\begin{tcolorbox}[breakable, colback=gray!5, colframe=gray!40, boxrule=0.5pt, sharp corners]
    \begin{lstlisting}[basicstyle=\ttfamily\small, breaklines=true, columns=fullflexible]
    id: hotpotqa_joint_facts_qa
name: HotpotQA Multi-hop QA with Supporting Facts (Distractor)
description: |
  Build a multi-hop QA system on the HotpotQA distractor setting that predicts both the final answer and the exact set of supporting sentences used to derive it. Each example provides a question, 10 candidate paragraphs (title, list of sentences), gold supporting facts as {{title, sent_id}} pairs, and an answer string {{which may be yes/no or a short span}}. You must use the provided sentence boundaries exactly {{do not re-split}} and select sentence indices from the given context. Train a multi-task model that: {{1}} predicts the answer, and {{2}} performs sentence-level classification over all candidate sentences to select supporting facts. Use either an extractive span head {{start/end indices with a separate yes/no head}} or a generative seq2seq model for the answer; use a classifier over sentence representations for supporting facts. Optimize a weighted sum of answer loss and supporting fact classification loss. Evaluate using the official HotpotQA metrics: Answer EM/F1, Supporting Facts EM/F1, and Joint EM/F1. Report Joint F1 as the primary metric, and include component metrics for analysis.
  Data and splits: - Use the train split for training and the validation/dev split for model selection and reporting. - This task uses the distractor setting {{10 provided paragraphs per question, where only a subset contains the supporting facts}}. - See {dataset_docs} for full dataset schema, features, and concrete examples.
  Modeling guidance: - Represent the 10-paragraph context with hierarchical encoders {{e.g., encode sentences, aggregate to paragraphs, then across paragraphs}}. - For long inputs, consider models that handle long sequences {{e.g., Longformer, BigBird}} or multi-hop retrieval to prune the context. - Supporting facts prediction is a multi-label classification over all candidate sentences in the provided context.
  Submission format: - You must submit a CSV file named submission.csv in the workspace root. - Columns:
    - id: the example id string {{e.g., "5a7a06935542990198eaf050"}}.
    - answer: the predicted answer string {{exact text; normalization is applied during evaluation}}.
    - supporting_facts: a JSON array of objects, each with keys "title" and "sent_id" indicating the selected supporting sentences.
  - Example lines:
    - id,answer,supporting_facts
    - 5a7a06935542990198eaf050,Arthur's Magazine,"[{{""title"": ""Arthur's Magazine"", ""sent_id"": 0}}, {{""title"": ""First for Women"", ""sent_id"": 0}}]"
    - 5a879ab05542996e4f30887e,Delhi,"[{{""title"": ""Oberoi family"", ""sent_id"": 0}}, {{""title"": ""The Oberoi Group"", ""sent_id"": 0}}]"
  - Constraints:
    - Each supporting fact must reference a title that exists in the example's context.title list and a valid sentence index {{0-based}} for that title.
    - No duplicates; order does not matter. Predict exactly the set of gold supporting sentences to achieve Supporting Facts EM.
    - Include as many sentences as required by the example {{often 2, but some questions require more}}.

  Evaluation and metrics: - Primary metric: Joint F1 {{combines answer F1 and supporting facts F1}}. - Secondary metrics reported: Answer EM/F1, Supporting Facts EM/F1, Joint EM. - Answer normalization follows HotpotQA conventions {{lowercasing, stripping punctuation and articles}}. - Supporting facts F1 computed over the set of {{title, sent_id}} pairs.
  Resources and expected runtime: - Training is expected to complete in ~3-5 hours on a single NVIDIA RTX A6000 with 8 CPU cores for pre-processing. - Keep memory usage in mind when encoding long contexts. Batch size and gradient accumulation may be required.
  Tips: - Start with a strong encoder {{e.g., DeBERTa-v3, RoBERTa}} with segment-level inputs and a sentence classification head. - Use curriculum: begin with answer-only training, then add supporting facts loss, or alternate batches. - Joint decoding heuristic: prioritize sentences from predicted relevant paragraphs; ensure coverage of all hops before final answer extraction/generation.
dataset_configs:
- datasets/hotpotqa_joint_facts_qa/hotpotqa_hotpot_qa.yaml
task_entrypoint: CSVSubmissionTasks
training_timeout: 18000
use_generic_conda: true
starter_code:
- data_train_v1/hotpotqa_joint_facts_qa/baseline.py
- data_train_v1/hotpotqa_joint_facts_qa/evaluate.py
baseline_paths:
- baseline.py
baseline_scores:
- joint_f1: 0.022210986997935424
  ans_em: 0.052532072923700206
  ans_f1: 0.08454953694131888
  sp_em: 0.008102633355840648
  sp_f1: 0.12714489351896444
  joint_em: 0.0013504388926401081
evaluation_paths:
- evaluate.py
evaluation_read_only: true
memory_path: data_train_v1/hotpotqa_joint_facts_qa/memory.json
    \end{lstlisting}
\end{tcolorbox}

\paragraph{baseline.py}
\begin{tcolorbox}[breakable, colback=gray!5, colframe=gray!40, boxrule=0.5pt, sharp corners]
    \begin{lstlisting}[basicstyle=\ttfamily\small, breaklines=true, columns=fullflexible]
    import csv
import json
from datasets import load_dataset
from tqdm import tqdm

YES_NO_STARTS = (
    "is", "are", "was", "were",
    "do", "does", "did",
    "can", "could", "may", "might", "must",
    "have", "has", "had",
    "will", "would", "should", "shall"
)

def simple_answer_heuristic(question, fallback_title):
    q = (question or "").strip().lower()
    if any(q.startswith(aux + " ") for aux in YES_NO_STARTS):
        return "yes"
    return fallback_title if fallback_title is not None else "unknown"

def main():
    # Use the distractor setting validation split
    ds = load_dataset("hotpotqa/hotpot_qa", "distractor", split="validation")

    with open("submission.csv", "w", newline="", encoding="utf-8") as f:
        writer = csv.DictWriter(f, fieldnames=["id", "answer", "supporting_facts"])
        writer.writeheader()

        for ex in tqdm(ds, desc="Generating baseline predictions"):
            ex_id = ex["id"]
            question = ex["question"]
            titles = ex["context"]["title"]
            sentences = ex["context"]["sentences"]

            # Choose up to two candidate supporting facts: first two titles with at least one sentence
            pred_sfs = []
            for idx, title in enumerate(titles):
                if idx < len(sentences) and len(sentences[idx]) > 0:
                    pred_sfs.append({"title": title, "sent_id": 0})
                if len(pred_sfs) == 2:
                    break

            # Fallbacks if not enough
            if not pred_sfs:
                # Ensure we still output something structurally valid
                if len(titles) > 0:
                    pred_sfs = [{"title": titles[0], "sent_id": 0}]
                else:
                    pred_sfs = []

            fallback_title = pred_sfs[0]["title"] if pred_sfs else (titles[0] if titles else None)
            answer = simple_answer_heuristic(question, fallback_title)

            writer.writerow({
                "id": ex_id,
                "answer": answer,
                "supporting_facts": json.dumps(pred_sfs, ensure_ascii=False)
            })

if __name__ == "__main__":
    main()
    \end{lstlisting}
\end{tcolorbox}

\paragraph{evaluate.py}
\begin{tcolorbox}[breakable, colback=gray!5, colframe=gray!40, boxrule=0.5pt, sharp corners]
    \begin{lstlisting}[basicstyle=\ttfamily\small, breaklines=true, columns=fullflexible]
import argparse
import csv
import json
import math
import re
import string
from collections import defaultdict
from datasets import load_dataset

PUNCT = set(string.punctuation)
ARTICLES = {"a", "an", "the"}
WHITESPACE_RE = re.compile(r"\s+")

def normalize_answer(s):
    if s is None:
        return ""
    s = s.lower()

    def remove_punc(text):
        return "".join(ch for ch in text if ch not in PUNCT)

    def remove_articles(text):
        return re.sub(r"\b(a|an|the)\b", " ", text)

    def white_space_fix(text):
        return WHITESPACE_RE.sub(" ", text).strip()

    return white_space_fix(remove_articles(remove_punc(s)))

def f1_score(prediction, ground_truth):
    pred_tokens = normalize_answer(prediction).split()
    gold_tokens = normalize_answer(ground_truth).split()
    if len(pred_tokens) == 0 and len(gold_tokens) == 0:
        return 1.0
    if len(pred_tokens) == 0 or len(gold_tokens) == 0:
        return 0.0
    common = defaultdict(int)
    for t in gold_tokens:
        common[t] += 1
    num_same = 0
    for t in pred_tokens:
        if common[t] > 0:
            num_same += 1
            common[t] -= 1
    if num_same == 0:
        return 0.0
    precision = num_same / len(pred_tokens)
    recall = num_same / len(gold_tokens)
    return 2 * precision * recall / (precision + recall)

def exact_match_score(prediction, ground_truth):
    return 1.0 if normalize_answer(prediction) == normalize_answer(ground_truth) else 0.0

def sp_em_f1(pred_set, gold_set):
    # pred_set and gold_set are sets of (title, sent_id) tuples
    inter = pred_set.intersection(gold_set)
    if len(gold_set) == 0 and len(pred_set) == 0:
        return 1.0, 1.0
    if len(gold_set) == 0:
        # No gold facts; treat as EM/F1 zero if pred non-empty; otherwise 1
        return (1.0 if len(pred_set) == 0 else 0.0), (1.0 if len(pred_set) == 0 else 0.0)
    em = 1.0 if pred_set == gold_set else 0.0
    if len(pred_set) == 0:
        return em, 0.0
    precision = len(inter) / len(pred_set)
    recall = len(inter) / len(gold_set)
    if precision + recall == 0:
        f1 = 0.0
    else:
        f1 = 2 * precision * recall / (precision + recall)
    return em, f1

def parse_supporting_facts(cell):
    try:
        data = json.loads(cell)
        out = set()
        if isinstance(data, list):
            for item in data:
                if isinstance(item, dict):
                    title = item.get("title", "")
                    sent_id = item.get("sent_id", 0)
                    try:
                        sent_id = int(sent_id)
                    except Exception:
                        # if cannot parse, skip this item
                        continue
                    out.add((title, sent_id))
        return out
    except Exception:
        return set()

def load_predictions_csv(path):
    preds = {}
    with open(path, "r", encoding="utf-8") as f:
        reader = csv.DictReader(f)
        for row in reader:
            ex_id = row.get("id", "")
            answer = row.get("answer", "")
            sf_cell = row.get("supporting_facts", "[]")
            pred_sfs = parse_supporting_facts(sf_cell)
            preds[ex_id] = {"answer": answer, "supporting_facts": pred_sfs}
    return preds

def main():
    parser = argparse.ArgumentParser()
    parser.add_argument("--submission_file", type=str, required=True)
    args = parser.parse_args()

    # Load dev split of HotpotQA distractor
    ds = load_dataset("hotpotqa/hotpot_qa", "distractor", split="validation")

    gold_by_id = {}
    for ex in ds:
        ex_id = ex["id"]
        answer = ex["answer"]
        titles = ex["supporting_facts"]["title"]
        sent_ids = ex["supporting_facts"]["sent_id"]
        gold_sfs = set()
        for t, s in zip(titles, sent_ids):
            try:
                s = int(s)
            except Exception:
                continue
            gold_sfs.add((t, s))
        gold_by_id[ex_id] = {"answer": answer, "supporting_facts": gold_sfs}

    preds = load_predictions_csv(args.submission_file)

    total = len(gold_by_id)
    ans_em_sum = 0.0
    ans_f1_sum = 0.0
    sp_em_sum = 0.0
    sp_f1_sum = 0.0
    joint_em_sum = 0.0
    joint_f1_sum = 0.0

    for ex_id, gold in gold_by_id.items():
        pred = preds.get(ex_id, {"answer": "", "supporting_facts": set()})

        a_em = exact_match_score(pred["answer"], gold["answer"])
        a_f1 = f1_score(pred["answer"], gold["answer"])
        s_em, s_f1 = sp_em_f1(pred["supporting_facts"], gold["supporting_facts"])

        ans_em_sum += a_em
        ans_f1_sum += a_f1
        sp_em_sum += s_em
        sp_f1_sum += s_f1
        joint_em_sum += (a_em * s_em)
        joint_f1_sum += (a_f1 * s_f1)

    metrics = {
        "joint_f1": joint_f1_sum / total if total > 0 else 0.0,
        "ans_em": ans_em_sum / total if total > 0 else 0.0,
        "ans_f1": ans_f1_sum / total if total > 0 else 0.0,
        "sp_em": sp_em_sum / total if total > 0 else 0.0,
        "sp_f1": sp_f1_sum / total if total > 0 else 0.0,
        "joint_em": joint_em_sum / total if total > 0 else 0.0,
    }
    # Print a single JSON object
    print(json.dumps(metrics))

if __name__ == "__main__":
    main()
    \end{lstlisting}
\end{tcolorbox}

\end{document}